\documentclass[10pt,twocolumn,letterpaper]{article}

\usepackage{cvpr}              

\def\cvprPaperID{*****} 
\def\confName{CVPR}
\def\confYear{2022}

\usepackage{cvpr}
\usepackage{booktabs}
\usepackage{times}
\usepackage{epsfig}
\usepackage{graphicx}
\usepackage{amsmath}
\usepackage{amssymb}
\usepackage{array}
\usepackage{multirow}
\usepackage[font=small,labelfont=bf]{caption}
\usepackage{xcolor}
\usepackage{color, colortbl}
\usepackage{cite}
\usepackage{enumitem}            
\usepackage{array, makecell} %
\usepackage{caption}
\usepackage{subcaption}

\usepackage{xcolor,pifont}
\newcommand*\colourcheck[1]{%
  \expandafter\newcommand\csname #1check\endcsname{\textcolor{#1}{\ding{52}}}%
}

\newcommand{\vect}[1]{\boldsymbol{#1}}

\newcommand*\colourcross[1]{%
  \expandafter\newcommand\csname #1check\endcsname{\textcolor{#1}{\ding{55}}}%
}

\setlist[itemize]{noitemsep,topsep=0pt,leftmargin=*,label={\large\textbullet}}

\colourcheck{green}
\colourcross{red}
\newcolumntype{P}[1]{>{\centering\arraybackslash}p{#1}}

\newcommand{\minisection}[1]{\vspace{0.7mm} \noindent \textbf{#1} \hspace{0mm}}

\newcommand{\Paragraph}[1]{\vspace{1mm} \noindent \textbf{#1} \hspace{0mm}}
\newcommand{\Section}[1]{\vspace{-1mm} \section{#1} \vspace{0mm}}
\newcommand{\SubSection}[1]{\vspace{-1mm} \subsection{#1} \vspace{0mm}}
\newcommand{\SubSubSection}[1]{\vspace{-1mm} \subsubsection{#1} \vspace{0mm}}

\def\cvprPaperID{6546} 

\usepackage[pagebackref,breaklinks,colorlinks]{hyperref}

\usepackage[capitalize]{cleveref}
\crefname{section}{Sec.}{Secs.}
\Crefname{section}{Section}{Sections}
\Crefname{table}{Table}{Tables}
\crefname{table}{Tab.}{Tabs.}

\begin{document}

\title{Proactive Image Manipulation Detection}

\author{
Vishal Asnani$^{1}$, \,\,
Xi Yin$^{2}$, \,\,
Tal Hassner$^{2}$, \,\,
Sijia Liu$^{1}$, \,\,
Xiaoming Liu$^{1}$ \\
$^1$Michigan State University,\,\,$^2$Meta AI\\
{\tt\small$^1$\{asnanivi, liusiji5, liuxm\}@msu.edu,\,\,$^2$\{yinxi, thassner\}@fb.com}}
\maketitle

\begin{abstract}
   Image manipulation detection algorithms are often trained to discriminate between images manipulated with particular Generative Models (GMs) and genuine/real images, yet generalize poorly to images manipulated with GMs unseen in the training. 
   Conventional detection algorithms receive an input image passively. By contrast,
   we propose a proactive scheme to image manipulation detection.   
   Our key enabling technique is to
   estimate a set of templates which when added onto the real image would lead to more accurate manipulation detection. 
   That is, a template protected real image, and its manipulated version, is better discriminated compared to the original real image vs.~its manipulated one.
   These templates are estimated using certain constraints based on the desired properties of templates. 
  For image manipulation detection, our proposed approach outperforms the prior work by an average precision of $16\%$ for CycleGAN and $32\%$ for GauGAN. 
 Our approach is generalizable to 
   a variety of
   GMs showing an improvement over prior work by an average precision of $10\%$ averaged across $12$ GMs. 
   Our code is available at \url{https://www.github.com/vishal3477/proactive_IMD}.
\end{abstract}

\Section{Introduction}

It's common for people to share personal photos on social networks. 
Recent developments of image manipulation techniques via Generative Models (GMs)~\cite{goodfellow2014generative} result in serious concerns over the authenticity of the images. 
As these techniques are easily accessible~\cite{disentangled-representation-learning-gan-for-pose-invariant-face-recognition,liu2019stgan, choi2018stargan, park2019gaugan, CycleGAN2017, stargan2,most-gan-3d-morphable-stylegan-for-disentangled-face-image-manipulation}, the shared images are at a greater risk for misuse after manipulation. Generation of fake images can be categorized into two types: entire image generation and partial image manipulation~\cite{wang2021faketagger, wang2020cnn}. 
While the former generates entirely new images by feeding a noise code to the GM, the latter involves the partial manipulation of a real image.
Since the latter alters the semantics of real images, it is generally considered as a greater risk, and thus partial image manipulation detection is the focus of this work.

Detecting such manipulation is an important step to alleviate societal concerns on the authenticity of shared images. 
Prior works have been proposed to combat manipulated media~\cite{unified-detection-of-digital-and-physical-face-attacks}. 
They leverage properties that are prone to being manipulated, including mouth movement~\cite{rossler2019faceforensics}, steganalysis features~\cite{wu2020sstnet}, attention mechanism~\cite{dang2020detection,pscc-net-progressive-spatio-channel-correlation-network-for-image-manipulation-detection-and-localization}, \etc.  
However, these methods are often overfitted to the image manipulation method and the dataset used in training, and suffer when tested on data with a different distribution.

\begin{figure}[t!]
\centering
\includegraphics[trim={0 -4 0 0},clip,width=\columnwidth]{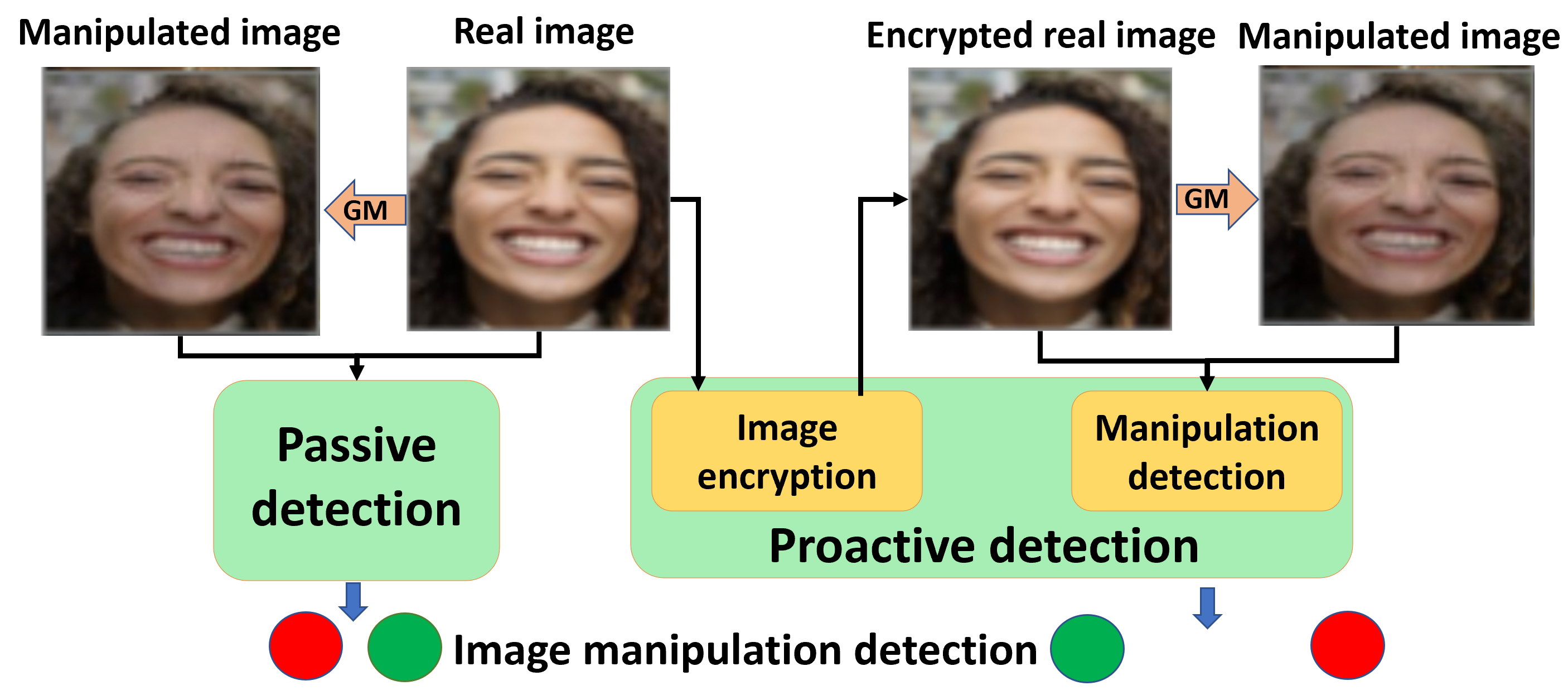}
\vspace{1mm}
\caption{\normalsize \textbf{Passive \vs proactive image manipulation detection} Classic passive schemes take an image as it is to discriminate a real image \vs its manipulated one created by a Generative Model (GM). In contrast, our proactive scheme performs encryption of the real image so that our detection module can better discriminate the encrypted real image \vs its manipulated counterpart.}
\label{fig:teaser}
\vspace{1mm}
\end{figure}

\begin{table*}[t]
\centering
\caption{\normalsize Comparison of our approach with prior works. Generalizable column means if the performance is reported on datasets unseen during training. [Keys: Img. man. det.: Image manipulation detection, Img. ind.: Image independent]}
\vspace{1.5mm}
\scalebox{0.68}{
\begin{tabular}{c|c|c|c|c|c|c|c|c|c|c}
\hline
\multirow{2}{*}{Method} & \multirow{2}{*}{Year} & Detection & \multirow{2}{*}{Purpose} & Manipulation & \multirow{2}{*}{Generalizable}  & Add & Recover & Template & \# of & Img. ind.\\
& & scheme & & type & & perturbation & perturbation & learning method & templates & templates\\\hline
Cozzolino~\etal~\cite{cozzolino2018forensictransfer} & $2018$& Passive & Img. man. det.  & Entire/Partial & \greencheck& \redcheck& \redcheck&-&-&-\\
Nataraj~\etal~\cite{nataraj2019detecting} & $2019$& Passive & Img. man. det.  & Entire/Partial & \greencheck& \redcheck& \redcheck&-&-&-\\
Rossler~\etal~\cite{rossler2019faceforensics} &$2019$& Passive & Img. man. det. & Entire/Partial & \redcheck& \redcheck& \redcheck&-&-&-\\
Zhang~\etal~\cite{zhang2019detecting} &$2019$& Passive & Img. man. det. & Partial & \greencheck&\redcheck& \redcheck&-&-&-\\
Wang~\etal~\cite{wang2020cnn} & $2020$& Passive & Img. man. det.  & Entire/Partial & \greencheck& \redcheck& \redcheck&-&-&-\\
Wu~\etal~\cite{wu2020sstnet} &$2020$& Passive & Img. man. det. & Entire/Partial & \redcheck& \redcheck& \redcheck&-&-&-\\
Qian~\etal~\cite{qian2020thinking} & $2020$& Passive & Img. man. det. & Entire/Partial& \redcheck& \redcheck& \redcheck&-&-&-\\
Dang~\etal~\cite{dang2020detection} &$2020$& Passive & Img. man. det. & Partial & \redcheck& \redcheck& \redcheck&-&-&-\\
Masi~\etal~\cite{masi2020two} &$2020$& Passive & Img. man. det. & Partial & \redcheck& \redcheck& \redcheck&-&-&-\\
Nirkin~\etal~\cite{nirkin2021deepfake} &$2021$& Passive & Img. man. det. & Partial & \redcheck& \redcheck& \redcheck&-&-&-\\
Asnani~\etal~\cite{asnani2021reverse} & $2021$& Passive & Img. man. det.  & Entire/Partial & \greencheck& \redcheck& \redcheck&-&-&-\\
Segalis~\etal~\cite{segalis2020ogan} &$2020$& Proactive & Deepfake disruption & Partial & \redcheck&\greencheck& \redcheck& Adversarial attack &$1$& \greencheck\\
Ruiz~\etal~\cite{ruiz2020disrupting} &$2020$& Proactive & Deepfake disruption & Partial & \redcheck&\greencheck& \redcheck& Adversarial attack&$1$& \greencheck\\
Yeh~\etal~\cite{yeh2020disrupting} &$2020$& Proactive & Deepfake disruption & Partial & \redcheck&\greencheck& \redcheck& Adversarial attack&$1$& \greencheck\\
Wang~\etal~\cite{wang2021faketagger} &$2021$& Proactive & Deepfake tagging & Partial & \redcheck&\greencheck& \greencheck& Fixed template &$>1$& \redcheck\\
\hline
Ours &-& Proactive & Img. man. det. & Partial & \greencheck&\greencheck& \greencheck& Unsupervised learning &$>1$& \greencheck\\
 \hline \hline
\end{tabular}}
\label{tab:rel_works}
\end{table*}

All the aforementioned methods adopt a {\it passive} scheme since the input image, being real or manipulated, is accepted as is for detection.
Alternatively, there is also a  {\it proactive} scheme proposed for a few computer vision tasks, which involves adding signals to the original image.
For example, prior works add a predefined template to real images which either disrupt the output of the GM~\cite{ruiz2020disrupting, yeh2020disrupting, segalis2020ogan} or  tag images to real identities~\cite{wang2021faketagger}. This template is either a one-hot encoding~\cite{wang2021faketagger} or an adversarial perturbation~\cite{ruiz2020disrupting, yeh2020disrupting, segalis2020ogan}.

Motivated by improving the generalization of manipulation detection, as well as the proactive scheme for other tasks, this paper proposes a {\it proactive} scheme for the purpose of image manipulation detection, which works as follows.
When an image is captured, our algorithm adds an imperceptible signal (termed as \textit{template}) to it, serving as an encryption.
If this encrypted image is shared and manipulated through a GM, our algorithm accurately distinguishes between the encrypted image and its manipulated version by recovering the added template.
Ideally, this encryption process could be incorporated into the camera hardware to protect all images after being captured.
In comparison, our approach differs from related proactive works~\cite{ruiz2020disrupting, yeh2020disrupting, segalis2020ogan,wang2021faketagger} in its purpose (detection vs other tasks), template learning (learnable vs predefined), the number of templates, and the generalization ability.

Our key enabling technique is to learn a template set, which is a non-trivial task.
First, there is no ground truth template for supervision. 
Second, recovering the template from manipulated images is challenging. 
Third, using one template can be risky as the attackers may reverse engineer the template.
Lastly, image editing operations such as blurring or compression could be applied to encrypted images, diminishing the efficacy of the added template.

To overcome these challenges, we propose a template estimation framework to learn a set of {\it orthogonal templates}. 
We perform image manipulation detection based on the recovery of the template from encrypted real and manipulated images. 
Unlike prior works, we use unsupervised learning to estimate  this template set based on certain constraints. 
We define different loss functions to incorporate properties including small magnitude, more high frequency content, orthogonality and classification ability as constraints to learn the template set. 
We show that our framework achieves superior manipulation detection than State-of-The-Art (SoTA) 
methods~\cite{wang2021faketagger,zhang2019detecting, cozzolino2018forensictransfer, nataraj2019detecting}. 
We propose a novel evaluation protocol with $12$ different GMs, where we train on images manipulated by one GM and test on unseen GMs. 
In summary, the contributions of this paper include:

\begin{itemize}
    \item We propose a novel proactive scheme for image manipulation detection. 
    \item We propose to learn a  set  of templates with desired properties, achieving higher performance than a single template approach.
    \item Our method substantially outperforms the prior works on image manipulation detection. 
     Our method is more generalizable to different GMs showing an improvement of $10\%$ average precision averaged across $12$ GMs.
\end{itemize}

\Section{Related Works}
\minisection{Passive deepfake detection.} 
Most deepfake detection methods are passive. 
Wang~\etal~\cite{wang2020cnn} perform binary detection by exploring frequency domain patterns from images. 
Zhang~\etal~\cite{zhang2019detecting} propose to extract the median and high frequencies to detect the upsampling artifacts by GANs.
Asnani~\etal~\cite{asnani2021reverse} propose to estimate fingerprint using certain desired properties for generative models which produce fake images.
Others use autoencoders ~\cite{cozzolino2018forensictransfer}, hand-crafted features~\cite{nataraj2019detecting}, face-context discrepancies~\cite{nirkin2021deepfake}, mouth and face motion~\cite{rossler2019faceforensics}, steganalysis features~\cite{wu2020sstnet}, xception-net~\cite{chollet2017xception}, frequency domain~\cite{masi2020two} and attention mechanisms~\cite{dang2020detection}. 
These aforementioned passive deepfake detection methods suffer from generalization.
We propose a novel proactive scheme for manipulation detection, aiming to improve the generalization.

\begin{figure*}[t!]
\centering
\includegraphics[width=\textwidth]{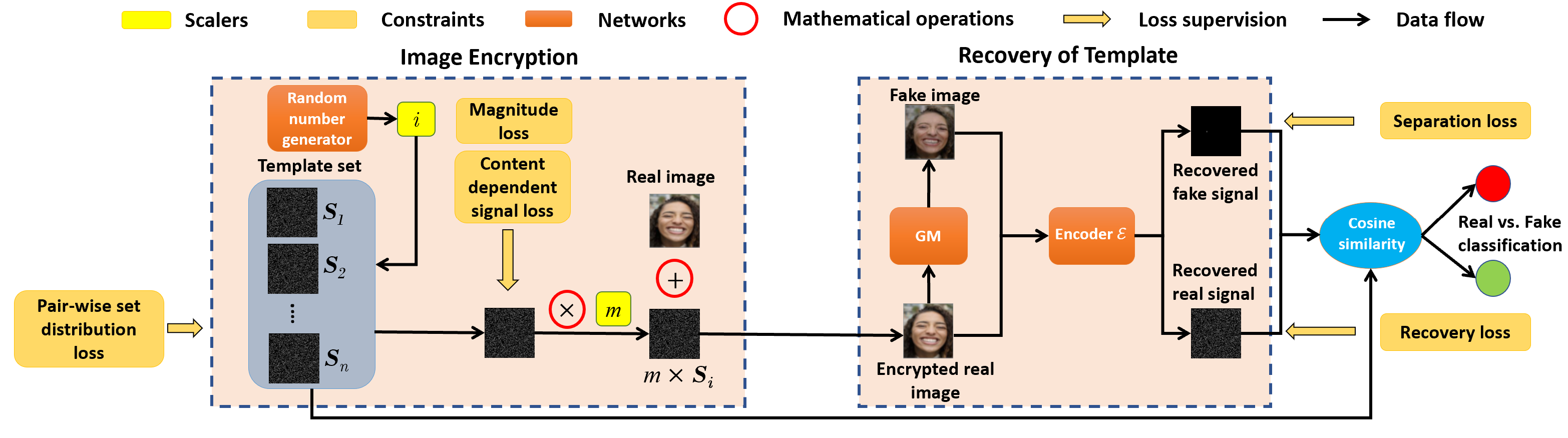}
\vspace{0.5mm}
\caption{\normalsize Our proposed framework includes two stages: 1) selection and addition of templates; and 2) the recovery of the estimated template from  encrypted real images and manipulated images using an encoder network. The GM is used in the inference mode. Both stages are trained in an end-to-end manner to output a set of templates. For inferences, the first stage is mandatory to encrypt the images. The second stage is used only when there is a need of image manipulation detection.}
\label{fig:overview}
\vspace{-3mm}
\end{figure*}

\minisection{Proactive schemes.} 
Recently, some proactive methods are proposed by adding an adversarial noise onto the real image. Ruiz~\etal~\cite{ruiz2020disrupting}
perform deepfake disruption by using adversarial attack in image translation networks.
Yeh~\etal~\cite{yeh2020disrupting} disrupt deepfakes to low quality images by performing adversarial attacks on real images. Segalis~\etal~\cite{segalis2020ogan} disrupt manipulations related to face-swapping by adding small perturbations. 
Wang~\etal~\cite{wang2021faketagger} propose a method to tag images by embedding messages and recovering them after manipulation. 
Wang~\etal~\cite{wang2021faketagger} use a one-hot encoding message instead of adversarial perturbations. 
Compared with these works, our method focuses on image manipulation detection rather than deepfake disruption or deepfake tagging. 
Our method learns a set of templates and recovers the added template for image manipulation detection. 
Our method also generalizes better to unseen GMs than prior works.
Tab.~\ref{tab:rel_works} summarizes the comparison  with prior works.

\minisection{Watermarking and cryptography methods.}
Digital watermarking methods have been evolving from using classic image transformation techniques to deep learning techniques. 
Prior work have explored different ways to embed watermarks through pixel values~\cite{bamatraf2010digital} and spatial domain~\cite{singh2012novel}.
Others~\cite{jiansheng2009digital, khan2013digital,yavuz2007improved} use frequency domains including transformation coefficients obtained via SVD, discrete wavelet transform (DWT), discrete cosine transform (DCT) and discrete fourier transform (DFT) to embed watermarks. Recently, deep learning techniques proposed by Zhu~\etal~\cite{zhu2018hidden}, Baluja~\etal~\cite{hide_image} and Tancik~\etal~\cite{tancik2020stegastamp} use an encoder-decoder architecture to embed watermarks into an image. 
All of these methods aim to either hide sensitive information or protect the ownership of digital images.
While our algorithm shares the high-level idea of image encryption, we develop a novel framework for an entirely different purpose, \ie, proactive image manipulation detection. 

\Section{Proposed Approach}
\SubSection{Problem Formulation}
\label{section:prob}
We only consider GMs which perform partial image manipulation that takes a real image as input for manipulation. Let $\vect{X}^a $ be a set of real images which when given as input to a GM $G$  would output $G(\vect{X}^a)$, a set of manipulated images.  Conventionally, passive image manipulation detection methods perform binary classification on $\vect{X}^a$ \vs $G(\vect{X}^a)$.  Denote $\vect{X}=\{\vect{X}^a, G(\vect{X}^a)\}$ $\in \mathbb{R}^{128\times128\times3}$ as the set of real and manipulated images, the objective function for passive detection is formulated as follows:

{\small
\vspace{-4mm}
\begin{equation}
   \min_{\theta}\bigg\{-\sum_j \Big (y_j. \text{log}(\mathcal{H}({\vect{X}_j};\theta))- \\
    (1-y_j).\text{log}(1-\mathcal{H}(\vect{X}_j;\theta))\Big ) \bigg\}.
    \label{eq:binary_obj}
\end{equation}
}%
where $y$ is the class label and $\mathcal{H}$ refers to the classification network used with parameters $\theta$. 

In contrast, for our proactive detection scheme, we apply a transformation  $\mathcal{T}$ to a real image from set $\vect{X}^a$ to formulate a set of encrypted real images represented as: $\mathcal{T}(\vect{X}^a)$. We perform image encryption by adding a learnable template to the image which acts as a defender's signature. 
Further, the set of encrypted real images $\mathcal{T}(\vect{X}^a)$ is given as input to the GM, which produces a set of manipulated images $G(\mathcal{T}(\vect{X}^a))$. 
We propose to learn a set of templates rather than a single one to increase security as it is difficult to reverse engineer all templates.
Thus for a real image $\vect{X}_j^a\in \vect{X}^a$, we define $\mathcal{T}$ via a set of {\it n} orthogonal templates ${\bf\mathcal{S}}=\{\vect{S}_1, \vect{S}_2,...\vect{S}_n\}$ where $\vect{S}_i$ $\in \mathbb{R}^{128\times128}$ as follows: 
\begin{equation}
    \mathcal{T}({\vect{X}_j^a})={\vect{X}_j^a} + {\vect{S}_i},  \text{ where }i\in \{1,2,...,n\}.
    \label{eq:add_temp}
\end{equation}

After applying the transformation $\mathcal{T}$, the objective function defined in Eqn.~\ref{eq:binary_obj} can be re-written as:
\begin{align}
    \begin{split}
    \min_{\theta,\vect{S}_i}&\bigg\{-\sum_j \Big (y_j. \text{log}(\mathcal{H}(\mathcal{T}(\vect{X}_j);\theta,\vect{S}_i))+\\
    &(1-y_j).\text{log}(1-\mathcal{H}(\mathcal{T}(\vect{X}_j);\theta,\vect{S}_i))\Big )\bigg\}.
    \end{split}
    \label{eq:bin_obj_with_x_tilde}
\end{align}

The goal is to find $\vect{S}_i$ for which corresponding images in $\vect{X}^a$ and $\mathcal{T}(\vect{X}^a)$ have no significant visual difference. 
More importantly, if $\mathcal{T}(\vect{X}^a)$ is modified by any GM, this would improve the performance for image manipulation detection. 

\setlength{\belowdisplayskip}{5pt}
\setlength{\abovedisplayskip}{5pt}
\captionsetup[sub]{font=footnotesize,labelfont={sf}}
\begin{figure}[t!]
\centering
\includegraphics[width=1\columnwidth]{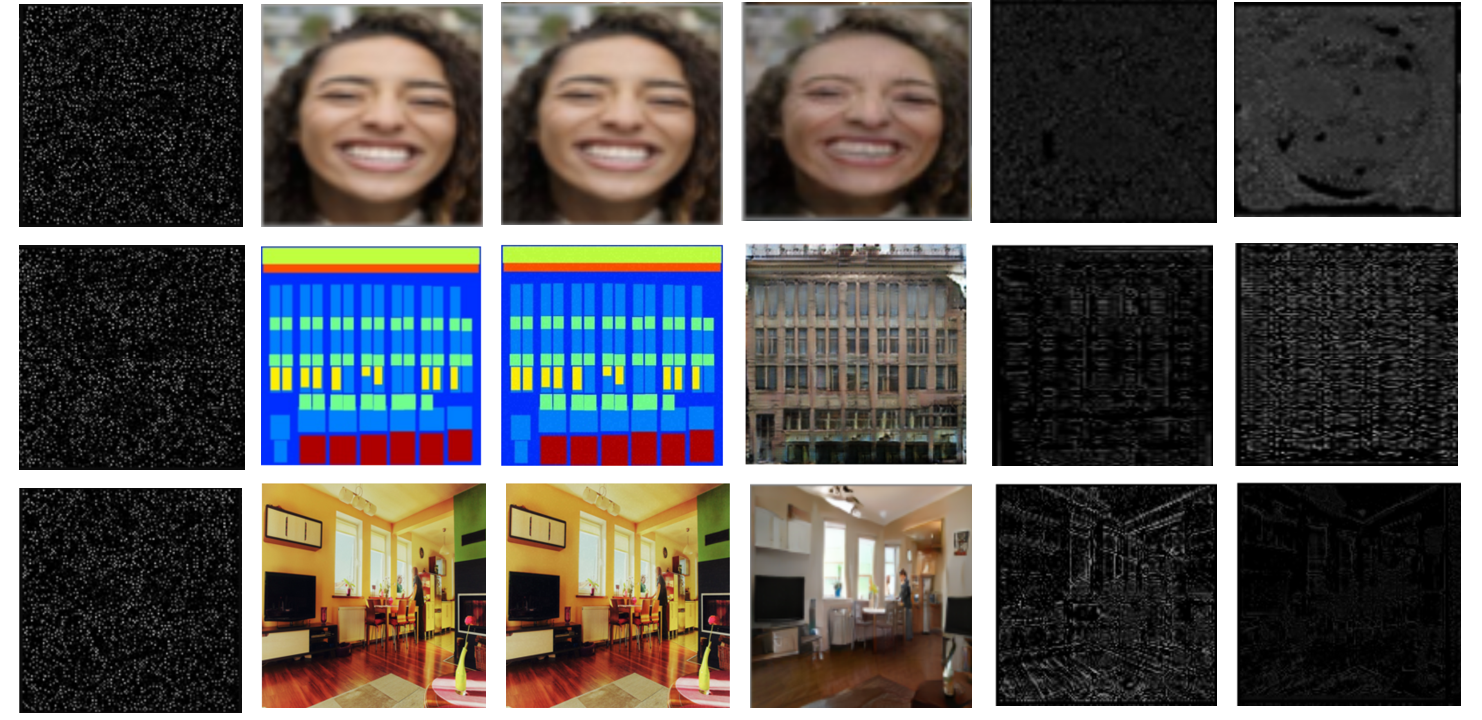}
\begin{minipage}[t]{.159\linewidth}
\vspace{-2mm} 
\centering
\subcaption{}
\end{minipage}%
\footnotesize {
\begin{minipage}[t]{.159\linewidth}
\vspace{-2mm} 
\centering
\subcaption{}
\end{minipage}
\begin{minipage}[t]{.159\linewidth}
\vspace{-2mm} 
\centering
\subcaption{}
\end{minipage}
\begin{minipage}[t]{.159\linewidth}
\vspace{-2mm} 
\centering
\subcaption{}
\end{minipage}
\begin{minipage}[t]{.159\linewidth}
\vspace{-2mm} 
\centering
\subcaption{}
\end{minipage}
\begin{minipage}[t]{.159\linewidth}
\vspace{-2mm} 
\centering
\subcaption{}
\end{minipage}}
\vspace{-1mm}
\caption{\normalsize Visualization of (a) a template set with the size of $3$, (b) real images, (c) encrypted real images after adding a template, (d) manipulated images output by a GM, (e) recovered template from (c), and (f) recovered template from (d). Each row corresponds to image manipulation by different GM (top: StarGAN, middle: CycleGAN, bottom: GauGAN). 
The template recovered from encrypted real images is more similar to the template set than the one from manipulated images. 
The addition of the template creates no visual difference between real and encrypted real images. We provide more examples of real images evaluated using our framework in the supplementary material.  }
\vspace{1mm}
\label{fig:sig_vis}
\end{figure}

\SubSection{Proposed Framework}
\label{section:framework}
As shown in Fig.~\ref{fig:overview}, our framework consists of two stages: image encryption and recovery of template. 
The first stage is used for selection and addition of templates, while the second stage involves the recovery of templates from images in  $\mathcal{T}(\vect{X}^a)$ and $G(\mathcal{T}(\vect{X}^a))$. 
Both stages are trained in an end-to-end manner with GM parameters fixed. 
For inference, each stage is applied separately. The first stage is a mandatory step to encrypt the real images while the second stage would only be used when image manipulation detection is needed. 

\SubSubSection{Image Encryption}
We initialize a set of $n$ templates as shown in Fig.~\ref{fig:overview}, which is optimized during training using certain constraints. 
As formulated in Eqn.~\ref{eq:add_temp}, we randomly select and add a template from our template set to every real image. 
Our objective is to estimate an optimal template set from which any template is capable of protecting the real image in $\vect{X}^a$.

Although we constrain the magnitude of the templates using $L_2$ loss, the added template still degrades the quality of the real image. Therefore, when adding the template to real images, we control the strength of the added template using a hyperparameter \textit{m}. 
We re-define $\mathcal{T}$ as follows:
\begin{equation}
    \mathcal{T}({\vect{X}^a_j})={\vect{X}^a_j} + m \times {\vect{S}_i} \text{ where}\hspace{1mm} i\in \{1,2,...,n\}.
\end{equation}

We perform an ablation study of varying \textit{m} in Sec. \ref{sec:abl_stu}, and find that setting \textit{m} at $30\%$ performs the best.

\SubSubSection{Recovery of Templates}
To perform image manipulation detection 
as shown in Fig.~\ref{fig:overview}, we attempt to recover our added template from images in $\mathcal{T}(\vect{X}^a)$ using an encoder $\mathcal{E}$ with parameters $\theta_{\mathcal{E}}$. 
For any real image $\vect{X}^a_j \in \vect{X}^a$, we define the recovered template from encrypted real image $\mathcal{T}(\vect{X}^a_j)$ as $\vect{S}_R=\mathcal{E}(\mathcal{T}({\vect{X}_j^a}))$ and from manipulated image $G(\mathcal{T}(\vect{X}^a_j))$ as $\vect{S}_F=\mathcal{E}(G(\mathcal{T}({\vect{X}_j^a})))$.
As template selection from the template set is random, 
the encoder receives more training pairs to learn how to recover any template from an image, which contributes positively to the robustness of the recovery process. We visualize our trained template set $\mathcal{S}$, and the recovered templates $\vect{S}_{R/F}$ in Fig. \ref{fig:sig_vis}.

The main intuition of our framework design is that $\vect{S}_R$ should be much more similar to the added template and vice-versa for $\vect{S}_F$. 
Thus, to perform image manipulation detection, we calculate the cosine similarity between $\vect{S}_{R/F}$ and all learned templates in the set $\mathcal{S}$ rather than merely using a classification objective. For every image, we select the maximum cosine similarity across all templates as the final score. Therefore, we update logit scores in Eqn.~\ref{eq:bin_obj_with_x_tilde} by cosine similarity scores as shown below: 
{\small
\begin{align}
\begin{split}
    &\min_{\theta_{\mathcal{E}},\vect{S}_i}\bigg\{\!-\!\sum_j \Big (y_j. \text{log}(\max_{i=1...n}(\text{Cos}(\mathcal{E}(\mathcal{T}({\vect{X}_j});\theta_{\mathcal{E}}), \vect{S}_i)))+\\
    &(1\!-\!y_j).\text{log}(1\!-\!\max_{i=1...n}(\text{Cos}(\mathcal{E}(\mathcal{T}({\vect{X}_j});\theta_{\mathcal{E}}), \vect{S}_i)))\Big )\bigg\}.
\end{split}
    \label{eqn:metric}
\end{align}
}%

\SubSubSection{Unsupervised Training of Template Set}
Since there is no ground truth for supervision, we define various constraints to guide the learning process. Let $\vect{S}$ be the template selected from set $\mathcal{S}$ to be added onto a real image. We formulate five loss functions as shown below.  

\minisection{Magnitude loss.} The real image and the encrypted image should be as similar as possible visually as the user does not want the image quality to deteriorate after template addition. 
Therefore, we propose the first constraint to regularize the magnitude of the template:
\begin{equation}
J_m = ||{\vect{S}}||_2^2. 
\label{eq:Jm}
\end{equation}

\minisection{Recovery loss.}We use an encoder network to recover the added template. 
Ideally, the encoder output, \ie, the recovered template $\vect{S}_R$ of the encrypted real image, should be the same as the original added template $\vect{S}$.
Thus, we propose to maximize the cosine similarity between these two templates:
\begin{equation}
J_r = 1-\text{Cos}(\vect{S},\vect{S}_R).
\label{eq:Jr}
\end{equation}

\minisection{Content independent template loss.} Our main aim is to learn a set of universal templates which can be used for detecting manipulated images from unseen GMs. These templates, despite being trained on one dataset, can be applied to images from a different domain. 
Therefore, we encourage the high frequency information in the template to be data independent. We propose a constraint to minimize low frequency information:  
\begin{equation}
J_c = ||\mathcal{L}(\mathbb{F}({\vect{S}}),k)||_2^2,
\label{eq:Jc}
\end{equation}
where $\mathcal{L}$ is the low pass filter selecting the $k \times k$ region in the center of the $2$D Fourier spectrum, while assigning the high frequency region to zero. $\mathbb{F}$ is the Fourier transform. 

\minisection{Separation loss.} We want the recovered template $\vect{S}_F$ from manipulated images $G(\mathcal{T}(\vect{X}))$ to be different than all the templates in set $\mathcal{S}$. Thus, we optimize $\vect{S}_F$ to be orthogonal to all the templates in the set $\mathcal{S}$. Therefore, we take the template for which the cosine similarity between $\vect{S}_F$ and the template is maximum, and minimize its respective cosine similarity:
\begin{equation}
J_s = \max_{i=1...n}(\text{Cos}(\mathcal{N}(\vect{S}_i),\mathcal{N}(\vect{S}_F))),
\label{eq:Js}
\end{equation}
where $\mathcal{N}(\vect{S})$ is the normalizing function defined as $\mathcal{N}(\vect{S})=(\vect{S}-\min(\vect{S}))/(\max(\vect{S})-\min(\vect{S}))$. 
Since this loss minimizes the cosine similarity to be $0$, we normalize the templates before similarity calculation.

\minisection{Pair-wise set distribution loss.}A template set would ensure that if the attacker is somehow able to get access to some of the templates, it would still be difficult to reverse engineer other templates. Therefore, we propose a constraint to minimize the inter-template cosine similarity to prompt the diversity of the templates in $\mathcal{S}$:
\begin{equation}
J_p = \mathop{\sum_{i=1}^{n}\sum_{j=i+1}^{n}}\text{Cos}(\mathcal{N}(\vect{S}_i),\mathcal{N}(\vect{S}_j)).\label{eq:Jp}
\end{equation}

The overall loss function for template estimation is thus:
\begin{equation}
J = \lambda_1{J_m}+\lambda_2{J_r}+\lambda_3{J_c}+\lambda_4{J_{s}}+\lambda_5{J_p},
\label{eqn:multi_temp}
\end{equation}
where $\lambda_1$, $\lambda_2$, $\lambda_3$, $\lambda_4$, $\lambda_5$ are the loss weights for each term. 
\Section{Experiments}
\vspace{1mm}
\SubSection{Settings}

\minisection{Experimental setup and dataset.}
We follow the experimental setting of Wang~\etal~\cite{wang2020cnn}, and  compare with four baselines:~\cite{wang2020cnn},~\cite{zhang2019detecting},~\cite{cozzolino2018forensictransfer} and~\cite{nataraj2019detecting}. 
For training,~\cite{wang2020cnn} uses $720K$ images from which the manipulated images are generated by ProGAN~\cite{karras2018progressive}. 
However, as our method requires a GM to perform partial manipulation, we choose STGAN~\cite{liu2019stgan} in training as ProGAN synthesizes entire images. 
We use $24K$ images in CelebA-HQ~\cite{karras2018progressive} as the real images and pass them through STGAN to obtain manipulated images for training. 
For testing, we use $200$ real images and pass them through {\it unseen} GMs such as StarGAN~\cite{choi2018stargan}, GauGAN~\cite{park2019gaugan} and CycleGAN~\cite{CycleGAN2017}. The real images for testing GMs are chosen from the respective dataset they are trained on, \ie CelebA-HQ for StarGAN, Facades~\cite{CycleGAN2017} for CycleGAN, and COCO~\cite{coco} for GauGAN.

\begin{table}[t]
\large
\centering
\caption{\normalsize Performance comparison with prior works.} 
\vspace{1.5mm}
\scalebox{0.71}{
\begin{tabular}{l|c|c|c|c|c}
\hline
\multirow{2}{*}{Method} & \multirow{2}{*}{Train GM} & Set & \multicolumn{3}{c}{Test GM Average precision (\%)}\\\cline{4-6}
 & &size& CycleGAN & StarGAN & GauGAN\\\hline\hline
\cite{nataraj2019detecting} & CycleGAN & - & $\bf100$ & $88.20$ & $56.20$\\
\cite{cozzolino2018forensictransfer} & ProGAN &-& $77.20$ & $91.70$ & $83.30$\\
\cite{zhang2019detecting} & AutoGAN &-& $\bf100$ & $\bf100$ & $61.00$\\
\cite{wang2020cnn} & ProGAN &-& $84.00$ & $\bf100$ & $67.00$\\\hline
\multirow{6}{*}{Ours} & \multirow{2}{*}{STGAN} &$3$& $96.12$ & $\bf100$ & $91.62$\\
& &$20$& $99.66$ & $\bf100$ & $90.58$\\\cline{2-6}
& \multirow{2}{*}{AutoGAN} &$3$& $97.87$ & $97.89$ & $86.57$\\
& &$20$& $97.05$ & $97.18$ & $84.24$\\\cline{2-6}
\addlinespace[0.03cm]& \makecell{STGAN +\\ AutoGAN} &$3$& $\bf100$ & $\bf100$ & $\bf99.69$\\\hline\hline
\end{tabular}}
\label{tab:baseline_comp}
\end{table}

\begin{table}[t]
\large
\centering
\caption{\normalsize Performance comparison with Wang~\etal~\cite{wang2020cnn}.} 
\vspace{1.5mm}
\scalebox{0.71}{
\begin{tabular}{l|c|c|c|c}
\hline
\multirow{2}{*}{Method} & \multirow{2}{*}{Train GM} & \multicolumn{3}{c}{Test GM TDR (\%) at low FAR (0.5\%)}\\\cline{3-5}
 & & CycleGAN & StarGAN & GauGAN\\\hline\hline
\cite{wang2020cnn} & ProGAN & $55.98$ & $93.88$ & $37.14$\\\hline
Ours & STGAN & $\bf88.50$ & $\bf100.00$ & $\bf43.00$\\\hline\hline
\end{tabular}}
\label{tab:baseline_comp_tdr}
\end{table}

To further evaluate generalization ability of our approach, we use $12$ additional unseen GMs that have diverse network architectures and loss functions, and are trained on different datasets. 
We manipulate each of $200$ real images with these $12$ GMs which gives $2,400$ manipulated images. 
The real images are chosen from the dataset that the respective GM is trained on. 
The list of GMs and their training datasets are provided in the supplementary. 

\begin{table*}[t!]
\large
\centering
\caption{\normalsize Average precision of $12$ testing GMs when our method is trained on only STGAN. 
All the GMs have different architectures and are trained on diverse datasets. 
The average precision of almost all GMs are over $90\%$ showing the generalization ability of our method. }
\vspace{1.5mm}
\scalebox{0.61}{
\begin{tabular}{l|c|c|c|c|c|c|c|c|c|c|c|c||c}
\hline
\multirow{2}{*}{GM} & UNIT & MUNIT & StarGAN2 & BicycleGAN & CONT\_Enc.  & SEAN & ALAE & Pix2Pix & DualGAN & CouncilGAN & ESRGAN & GANimation & \multirow{2}{*}{Average}\\
&\cite{liu2018unit}&\cite{huang2018munit}&\cite{stargan2}&\cite{zhu2017bicycle}&\cite{pathakCVPR16context}&\cite{Zhu_2020_sean}&\cite{pidhorskyi2020alae}&\cite{isola2017pix2pix}&\cite{yi2018dualgan}&\cite{nizan2020council}&\cite{wang2021realesrgan}&\cite{Pumarola_ijcv2019gananime}&\\\hline
\cite{wang2020cnn} & $64.94$ & $95.33$ & $\bf100$  & $\bf 100$ & $98.18$ & $67.81$ & $92.73$ & $91.26$ & $\bf98.91$ & $74.13$ & $57.04$  & $55.19$ & $82.97$\\
 Ours & $\bf100$ & $\bf100$ & $\bf100$ & $99.05$ & $\bf98.75$ & $\bf97.63$ & $\bf93.10$ & $\bf92.50$ & $92.49$ & $\bf89.71$ & $\bf87.30$ & $\bf58.69$ & $\bf92.43$\\\hline\hline
\end{tabular}
}
\label{tab:gm_gen}
\end{table*}

\begin{table}[t]
\normalsize
\centering
\caption{\normalsize Performance comparison of our proposed method with Ruiz~\etal~\cite{ruiz2020disrupting}. The performance for our proposed method is better than~\cite{ruiz2020disrupting} when the testing GM is unseen. Both methods use StarGAN as the training GM.}
\vspace{1.5mm}
\scalebox{0.82}{
\begin{tabular}{l|c|c|c|c}
\hline
\multirow{2}{*}{Method} & \multicolumn{4}{c}{Test GM Average precision (\%)}\\\cline{2-5}
 &  StarGAN & CycleGAN & GANimation & Pix2Pix\\\hline\hline
\cite{ruiz2020disrupting}  & $\bf100$ & $51.50$& $52.43$& $49.08$\\
Ours  & $\bf100$ & $\bf95.26$ & $\bf60.12$& $\bf91.85$ \\
\hline\hline
\end{tabular}
}
\label{tab:baseline_comp_proactive}
\end{table}

\minisection{Implementation details.} 
Our framework is trained end-to-end for $10$ epochs via Adam optimizer with a learning rate of $10^{-5}$ and a batch size of $4$.
The loss weights are set to ensure similar magnitudes at the beginning of training: $\lambda_1=100$, $\lambda_2 = 30$, $\lambda_3 = 5$, $\lambda_4 = 0.003$, $\lambda_5 = 10$. 
If not specified, we set the template set size $n=3$. 
We set $k=50$ in the content independent template loss. 
All experiments are conducted using one NVIDIA Tesla K$80$ GPU.

\minisection{Evaluation metrics.} We 
report average precision as adopted by~\cite{wang2020cnn}. 
To mimic real-world scenarios, we further report true detection rate (TDR) at a low false alarm rate (FAR) of $0.5\%$.

\SubSection{Image Manipulation Detection Results}
\label{sec:imd}
As shown in Tab.~\ref{tab:baseline_comp}, when our training GM is STGAN, we can outperform the baselines by a large margin on GauGAN-based test data, while the performance on StarGAN-based test data remains the same at $100\%$. 
When training on STGAN, our method achieves lower performance on CycleGAN. We hypothesis that it is because AutoGAN and CycleGAN share the same model architecture. To validate this, we change our training GM to AutoGAN and observe improvement when tested on CycleGAN. However, the performance drops on other two GMs because the amount of training data is reduced ($24K$ for STGAN and $1.5K$ for AutoGAN). Increasing the number of templates can improve the performance for when trained on STGAN and test on CycleGAN, but degrades for others. The degradation is more when train on AutoGAN. It suggests that it is challenging to find a larger template set on a smaller training set. Finally, using both STGAN and AutoGAN training data can achieve the best performance.


\minisection{TDR at low FAR.} We also evaluate using TDR at low FAR in Tab.~\ref{tab:baseline_comp_tdr}. 
This is more indicative of the performance in the real world application where the number of real images are exponentially larger than manipulated images. 
For comparison, we evaluate the pretrained model of~\cite{wang2020cnn} on our test set. 
Our method performs consistently better for all three GMs, demonstrating the superiority of our approach.  

\minisection{Generalization ability.}
To test our generalization ability, 
we perform extensive evaluations across a large set of GMs. We compare the performance of our method with~\cite{wang2020cnn} by evaluating its pretrained model on a test set of different GMs.
Our framework performs quite well on almost all the GMs compared to~\cite{wang2020cnn} as shown in Tab.~\ref{tab:gm_gen}. This further demonstrates the generalization ability of our framework in the real world where an image can be manipulated by any unknown GM. Compared to \cite{wang2020cnn}, our framework achieves an improvement in the average precision of almost $10\%$ averaged across all $12$ GMs. 

\minisection{Comparison with proactive scheme work.} We compare our work with previous work in proactive scheme~\cite{ruiz2020disrupting}. 
As~\cite{ruiz2020disrupting} proposes to disrupt the GM's output, they only provide the distortion results of the manipulated image. 
To enable binary classification, we take their adversarial real and disrupted fake images to train a classifier with the similar network architecture as our encoder. 
Tab.~\ref{tab:baseline_comp_proactive} shows that~\cite{ruiz2020disrupting} works perfectly when the testing GM is the same as the training GM.
Yet if the testing GM is unseen, the performance drops substantially. 
Our method performs much better showing the high generalizability.

\minisection{Comparison with steganography works.}
Our method aligns with the high-level idea of digital steganographhy methods~\cite{bamatraf2010digital,singh2012novel,yavuz2007improved,Zhu_2020_sean, hide_image} which are used to hide an image onto other images. We compare our approach to the recent deep learning-based steganography method, Baluja~\etal~\cite{hide_image}, with its publicly available code. We hide and retrieve the template using the pre-trained model provided by~\cite{hide_image}. Our approach has far better average precision for each test GM compared to~\cite{hide_image} as shown in Tab.~\ref{tab:stega_aa_met}. This validates the effectiveness of template learning and concludes that the digital steganography methods are less generalizable across unknown GMs than our approach.

\minisection{Comparison with benign adversarial attacks.} Adversarial attacks are used to optimize a perturbation to change the class of the image. The learning of the template using our framework is similar to a benign usage of adversarial attacks. We conduct an ablation study to compare our method with common attacks such as benign PGD and FGSM. We remove the losses in Eqs.~\ref{eq:Jm},~\ref{eq:Jc}, and~\ref{eq:Jp} responsible for learning the template and replace them with an adversarial noise constraint. 
Our approach has better average precision for each test GM than both adversarial attacks as shown in Tab.~\ref{tab:stega_aa_met}. We observe that adversarial noise performed similar to passive schemes offering poor generalization to unknown GMs. This shows the importance of using our proposed constraints to learn the universal template set. 

\begin{table}[t]
\large
\centering
\caption{\normalsize Performance comparison of our proposed method with steganography and adversarial attack methods.} 
\vspace{1.5mm}
\scalebox{0.7}{
\begin{tabular}{l|c|c|c|c}
\hline
\multirow{2}{*}{Method} & \multirow{2}{*}{Type} & \multicolumn{3}{c}{Test GM Average precision (\%)}\\\cline{3-5}
 && CycleGAN & StarGAN & GauGAN\\\hline\hline
Baluja~\cite{hide_image} & Steganography & $85.64$ & $88.06$ & $81.26$\\\hline
PGD~\cite{madry2018towards} & Adversarial  & $90.28$ & $98.22$ & $57.71$\\
FGSM~\cite{goodfellow2014explaining} & attack & $89.21$ & $98.29$ & $63.81$\\\hline
Ours & - & $\bf99.95$ & $\bf100$ & $\bf98.23$\\
\hline\hline
\end{tabular}
}
\label{tab:stega_aa_met}
\end{table}

\minisection{Data augmentation.}
We apply various data augmentation schemes to evaluate the robustness of our method. 
We adopt some of the image editing techniques from Wang~\etal~\cite{wang2020cnn},  
including (1) Gaussian blurring, (2) JPEG compression, (3) blur $+$ JPEG ($0.5$), and (4) blur $+$ JPEG ($0.1$), where $0.5$ and $0.1$ are the probabilities of applying these image editing operations. 
In addition, we add resizing, cropping, and Gaussian noise. 
The implementation details of these techniques are in the supplementary. These techniques are applied after addition of our template to the real images.

\begin{table}[t]
\centering
\caption{\normalsize Average precision (\%) with various augmentation techniques in training and testing for three GMs.
We apply data augmentation to three scenarios: (1) in training only (2) in testing only and (3) in both training and testing. 
[Keys: aug.=augmentation, B.=blur, J.=JPEG compression, Gau.~No.=Gaussian Noise]}
\vspace{1.5mm}
\scalebox{0.69}{
\begin{tabular}{l|c|c|c|c|c|c}
\hline
\multicolumn{2}{c|}{Augmentation} & Augmentation & \multirow{2}{*}{Method} & \multicolumn{3}{c}{Test GMs}\\\cline{1-2} \cline{5-7}
Train & Test & type & & CycleGAN & StarGAN & GauGAN\\\hline
\multirow{2}{*}{\redcheck} & \multirow{2}{*}{\redcheck} & No &~\cite{wang2020cnn} & $84.00$ & $\bf100$ & $67.00$\\
& & augmentation & Ours & $96.12$ & $\bf100$ & $91.62$\\\hline

\multirow{11}{*}{\greencheck} & \multirow{11}{*}{\redcheck} & \multirow{2}{*}{Blur} &~\cite{wang2020cnn} & $90.10$ & $\bf100$ & $74.70$\\
& & & Ours & $93.55$ & $\bf100$ & $92.35$\\\cline{3-7}
& & \multirow{2}{*}{JPEG} &~\cite{wang2020cnn} & $93.20$ & $91.80$ & $97.50$\\
& & & Ours & $98.74$ & $98.30$ & $91.85$\\\cline{3-7}
& & \multirow{2}{*}{B$+$J ($0.5$)} &~\cite{wang2020cnn} & $96.80$ & $95.40$ & $98.10$\\
& & & Ours & $94.44$ & $\bf100$ & $98.16$\\\cline{3-7}
& & \multirow{2}{*}{B$+$J ($0.1$)} &~\cite{wang2020cnn} & $93.50$ & $84.50$ & $89.50$\\
& & & Ours & $95.79$ & $\bf100$ & $95.94$\\\cline{3-7}
& & Resizing & \multirow{3}{*}{Ours} & $\bf100$ & $\bf100$ & $98.97$\\
& & Crop & & $84.45$ & $84.92$ & $94.43$\\
& & Gau.~No. & & $99.95$ & $\bf100$ & $\bf99.11$\\\hline

\multirow{6}{*}{\redcheck} & \multirow{6}{*}{\greencheck} & Blur & \multirow{6}{*}{Ours} & $95.74$ & $84.87$ & $70.74$\\
& & JPEG &  & $91.91$ & $82.96$ & $84.16$\\
& & B$+$J ($0.5$) &  & $89.23$ & $82.18$ & $75.53$\\
& & Resizing &  & $93.12$ & $77.41$ & $91.45$\\
& & Crop &  & $84.04$ & $73.87$ & $70.12$\\
& & Gau.~No. &  & $73.83$ & $69.47$ & $66.70$\\\hline

\multirow{6}{*}{\greencheck} & \multirow{6}{*}{\greencheck} & Blur & \multirow{6}{*}{Ours} & $92.16$ & $\bf100$ & $90.15$\\
& & JPEG &  & $94.00$ & $97.92$ & $85.91$\\
& & B$+$J ($0.5$) &  & $87.37$ & $84.92$ & $74.68$\\
& & Resizing &  & $99.98$ & $\bf100$ & $92.73$\\
& & Cropping &  & $77.63$ & $89.22$ & $79.96$\\
& & Gau.~No. &  & $97.44$ & $\bf100$ & $82.32$\\

\hline\hline
\end{tabular}}
\label{tab:data_aug}
\end{table}

We evaluate in three scenarios when augmentation is applied in $(1)$ training, $(2)$ testing, $(3)$ both training and testing. As shown in Tab.~\ref{tab:data_aug}, for the augmentation techniques adopted from \cite{wang2020cnn}, we outperform~\cite{wang2020cnn} in almost all techniques. 
We observe significant improvement when blurring or JPEG compression is applied jointly but the  improvement is less when they are applied separately. 



As for the different scenarios on when data augmentation is applied, scenario $2$ performs the worst because the augmentation applied in testing has not been seen during training. Scenario $3$ performs better than scenario $2$ in most cases. There is a much larger performance drop when blurring and JPEG are applied together than separately.  Cropping performs the worst for both Scenario $1$ and $3$.

\begin{figure}[t!]
\centering
\includegraphics[width=1\columnwidth, height=36mm]{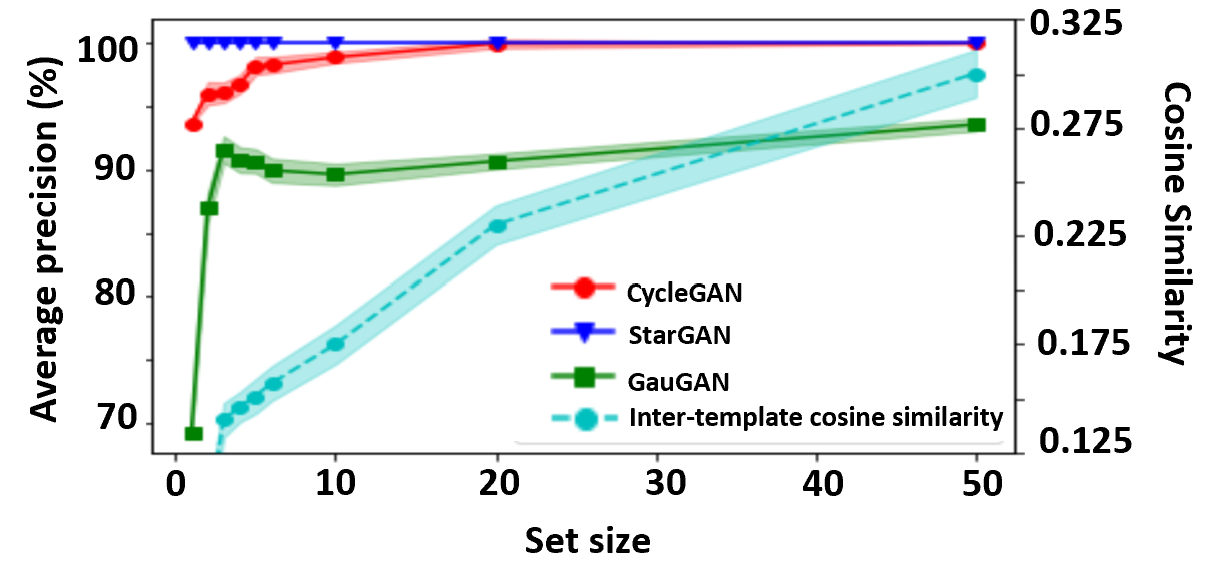}
\vspace{0.5mm}
\caption{\normalsize  Ablation study with varying template set sizes. The  performance improves when the set size increases, while the inter-template cosine similarity also increases. }
\vspace{0.5mm}
\label{fig:set_size}
\end{figure}

\SubSection{Ablation Studies}
\label{sec:abl_stu}
\minisection{Template set size.}
We study the effects of the template set size. 
As shown in Fig.~\ref{fig:set_size},  
the average precision increases as the set size is expanding from $1$ and saturates around the set size $10$. 
In the meantime, the average cosine similarity between templates within the set increases consistently, as it gets harder to find many orthogonal templates. We also test our framework's run-time for different set sizes. 
On a Tesla K$80$ GPU, for the set size of $1,3,10,20$ and $50$, the per-image run-time of our manipulation detection is $26.19$, $27.16$, $28.44$, $34.26$, and $43.76$ ms respectively. 
Thus, despite increasing the set size enhances our accuracy and security, there is a trade-off with the detection speed which is a important factor too. 
For comparison, we also test the pretrained model of~\cite{wang2020cnn} which gives a per-image run-time of $54.55$ ms. 
Our framework is much faster even with a larger set size which is due to the shallow network in our proactive scheme compared to a deeper network in passive scheme.

\begin{figure}[t!]
\centering
\includegraphics[width=1\columnwidth, height=36mm]{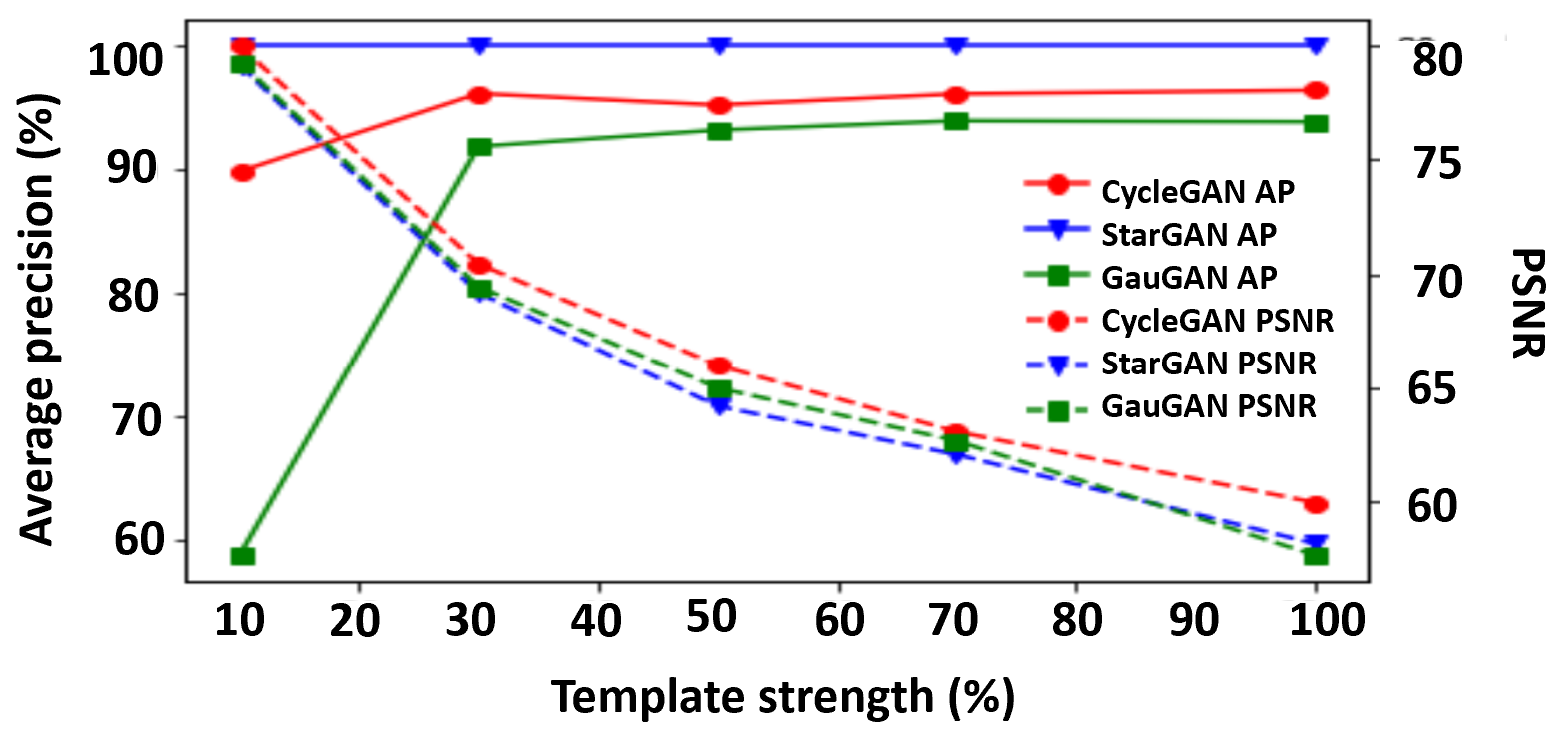}
\vspace{0.5mm}
\caption{\normalsize Ablation with varying template strengths in the encrypted real images. The lower the template strength, the higher the PSNR is and the harder it is for our encoder to recover it, which leads to lower detection performance.} 
\vspace{0.5mm}
\label{fig:sig_str}
\end{figure}

\minisection{Template strength.}
We use a hyperparameter $m$ to control the strength of our added template.
We ablate $m$ and show the results in Fig.~\ref{fig:sig_str}.
Intuitively, the lower the strength of the template added, the lower the detection performance since it would be harder for the encoder to recover the original template. 
Our results support this intuition. 
For all three GMs, the precision increases as we enlarge the template strength, and converges after $50\%$ strength.
We also show the PSNR between the encrypted real image and the original real image.
The PSNR decreases as we enlarge the strength as expected. 
We choose $m=30\%$ for a trade-off between the detection precision and the visual quality.

\begin{table}[t]
\centering
\caption{\normalsize Ablation study to remove losses used in our training. Removing any one loss deteriorates the performance compared to our proposed method. Fixing the template or performing direct classification made the results worse. This shows the importance of a variable template and using an encoder for classification purposes.}
\vspace{1.5mm}
\scalebox{0.73}{
\begin{tabular}{l|c|c|c}
\hline
\multirow{2}{*}{Loss removed} & \multicolumn{3}{c}{Test GM Average precision (\%)}\\\cline{2-4}
 & CycleGAN & StarGAN & GauGAN\\\hline\hline
Magnitude loss $(J_m)$ & $94.43$ & $\bf100$ & $87.44$\\
Pair-wise set distribution loss $(J_p)$ & $66.60$ & $79.99$ & $74.55$\\
Recovery loss $(J_r)$ & $51.59$ & $94.18$ & $90.61$\\
Content independent template loss $(J_c)$ & $92.01$ & $\bf100$ & $80.54$\\
Separation loss $(J_s)$ & $92.24$ & $\bf100$ & $64.06$\\
$J_m$, $J_p$ and $J_c$ (fixed template)&  $46.93$ & $59.88$ & $43.64$\\
$J_r$ and $J_s$ (removing encoder) & $50.00$ & $98.24$ & $55.00$\\
None (ours) & $\bf96.12$ & $\bf100$ & $\bf91.62$\\
\hline\hline
\end{tabular}}
\label{tab:loss_ablation}
\end{table}

\minisection{Loss functions.}
Our training process is guided by an objective function with five losses (Eqn.~\ref{eqn:multi_temp}).
To demonstrate the necessity of each loss, we ablate by removing each loss and compare with our full model. 
As shown in Tab.~\ref{tab:loss_ablation}, removing any one of the losses results in performance degradation. Specifically, removing the pair-wise set distribution loss, recovery loss or separation loss causes a larger drop. 

To better understand the importance of the data-driven template set, we fix the template set during training, \ie, removing the three losses directly operating on the template and only considering recovery and separation losses for training. 
We observe a significant performance drop, which shows that the learnable template is indeed crucial for effective image manipulation detection.

Finally, we remove the encoder from our framework and use a classification network with similar number of layers. 
Instead of recovering templates, the classification network is directly trained to perform binary image manipulation detection via cross-entropy loss. 
The performance drops significantly. 
This observation aligns with the previous works~\cite{wang2020fakespotter, cozzolino2018forensictransfer, zhang2019detecting} stating that CNN networks trained on images from one GM show poor generalizability to unseen GMs. 
The performance drops for all three GMs but CycleGAN and GauGAN are affected the most, as the datasets are different. 
For our proposed approach, when we are recovering the template, the encoder ignores all the low frequency information of the images which are data dependent. 
Thus, being more data (\ie, image content) independent, our encoder is able to achieve a higher generalizability.

\minisection{Template selection.}
Given a real image, we randomly select a template from the learnt template set to add to the image.
Thus, every image has an equal chance of selecting any one template from the set, resulting in many combinations for the entire test set. 
This raises the question of finding a worst and best combination of templates for all images in the test set.
To answer this, we experiment with a template set size of $50$ as a large size may offer higher variation in performance. 
For each image in $\mathcal{T}(\vect{X}^a)$ and $G(\mathcal{T}(\vect{X}^a))$, we calculate the cosine similarity between added template $\vect{S}$ and recovered template $\vect{S}_{R/F}$. For the worst/best case of every image, we select the template with the minimum/maximum difference between the real and manipulated image cosine similarities. 
As shown in Tab.~\ref{tab:best_worst}, GauGAN gives much more variation in the performance compared to CycleGAN and StarGAN. 
This shows that the template selection is an important step for image manipulation detection. 
This brings up the idea of training a network to select the best template for a specific image, by using the best case described above as a pseudo ground truth to supervise the network. 
We hypothesis template selection could be important, but with experiments, the difference of performance among different templates is nearly zero and the network's selection doesn't help in the performance compared with selecting the template randomly as shown in Tab.~\ref{tab:best_worst}. 
Therefore, we cannot have a pseudo ground truth to train another network for template selection. 

Another option for template selection is to select the {\it same} template for every test image which is equivalent to using one template compromising the security of our method. 
Nevertheless, we test this option to see the performance variation of biasing one template for all images. 
The performance variation is larger than our random selection scheme.
This shows that each template has a similar contribution to image manipulation detection.

\begin{table}[t]
\large
\centering
\caption{\normalsize Ablation of template selection schemes at set size of $50$. }
\vspace{1.5mm}
\scalebox{0.7}{
\begin{tabular}{l|c|c|c}
\hline
\multirow{2}{*}{Selection scheme} & \multicolumn{3}{c}{Test GM Average precision (\%)}\\\cline{2-4}
 & CycleGAN & StarGAN & GauGAN\\\hline\hline
Random selection & $99.90 \pm 0.02 $ & $\bf100 \pm 0.00 $ & $93.56 \pm 0.52 $\\
Biasing one template & $99.05 \pm 0.37$ & $\bf100 \pm 0.00 $ & $91.21 \pm 0.97$\\
Network based & $95.46 $ & $\bf 100 $ & $90.47$\\\hline
Worst case & $94.85$ & $\bf100$ & $80.55$\\
Best case & $\bf99.95$ & $\bf100$ & $\bf98.23$\\
\hline\hline
\end{tabular}}
\label{tab:best_worst}
\end{table}

\Section{Conclusion}
In this paper, we propose a proactive scheme for image manipulation detection. 
The main objective is to estimate a set of templates, which when added to the real images improves the performance for image manipulation detection. 
This template set is estimated using certain constraints and any template can be added onto the image right after it is being captured by any camera. Our framework is able to achieve better image manipulation detection performance on different unseen GMs, compared to prior works. 
We also show the results on a diverse set of $12$ additional GMs to demonstrate the generalizability of our proposed method. 

\Paragraph{Limitations.}
First,
although our work aims to protect real images in a proactive manner and can detect whether an image has been manipulated or not, it cannot perform general deepfake detection on entirely synthesized images.
Second, we try our best to collect a diverse set of GMs to validate the generalization of our approach. 
However, there are many other GMs that do not have open-sourced codes to be evaluated in our framework.
Lastly, how to supervise the training of a network for template selection is still an unanswered question.   

\Paragraph{Potential societal impact.}
We propose a proactive scheme which uses encrypted real images and their manipulated versions to perform manipulation detection. 
While this offers more generalizable detection, the encrypted real images might be used for training GMs in the future, which could make the manipulated images more robust against our framework, and thus warrents more research. 

\nocite{*}
{\small
\bibliographystyle{ieee_fullname}
\bibliography{egbib}

\begin{thebibliography}{10}\itemsep=-1pt

\bibitem{afchar2018mesonet}
Darius Afchar, Vincent Nozick, Junichi Yamagishi, and Isao Echizen.
\newblock Meso{N}et: a compact facial video forgery detection network.
\newblock In {\em WIFS}, 2018.

\bibitem{Agustsson_2017_CVPR_Workshops}
Eirikur Agustsson and Radu Timofte.
\newblock {NTIRE} 2017 challenge on single image super-resolution: Dataset and
  study.
\newblock In {\em CVPRW}, 2017.

\bibitem{asnani2021reverse}
Vishal Asnani, Xi Yin, Tal Hassner, and Xiaoming Liu.
\newblock Reverse engineering of generative models: Inferring model
  hyperparameters from generated images.
\newblock {\em arXiv preprint arXiv:2106.07873}, 2021.

\bibitem{hide_image}
Shumeet Baluja.
\newblock Hiding images in plain sight: Deep steganography.
\newblock 2017.

\bibitem{bamatraf2010digital}
Abdullah Bamatraf, Rosziati Ibrahim, and Mohd Najib B~Mohd Salleh.
\newblock Digital watermarking algorithm using {LSB}.
\newblock In {\em ICCAIE}, 2010.

\bibitem{coco}
Holger Caesar, Jasper Uijlings, and Vittorio Ferrari.
\newblock Coco-stuff: Thing and stuff classes in context.
\newblock In {\em CVPR}, 2018.

\bibitem{choi2018stargan}
Yunjey Choi, Minje Choi, Munyoung Kim, Jung-Woo Ha, Sunghun Kim, and Jaegul
  Choo.
\newblock Star{GAN}: Unified generative adversarial networks for multi-domain
  image-to-image translation.
\newblock In {\em CVPR}, 2018.

\bibitem{stargan2}
Yunjey Choi, Youngjung Uh, Jaejun Yoo, and Jung-Woo Ha.
\newblock Star{GAN} v2: Diverse image synthesis for multiple domains.
\newblock In {\em CVPR}, 2020.

\bibitem{chollet2017xception}
Fran{\c{c}}ois Chollet.
\newblock Xception: Deep learning with depthwise separable convolutions.
\newblock In {\em CVPR}, 2017.

\bibitem{cozzolino2018forensictransfer}
Davide Cozzolino, Justus Thies, Andreas R{\"o}ssler, Christian Riess, Matthias
  Nie{\ss}ner, and Luisa Verdoliva.
\newblock Forensictransfer: Weakly-supervised domain adaptation for forgery
  detection.
\newblock {\em arXiv preprint arXiv:1812.02510}, 2018.

\bibitem{dang2020detection}
Hao Dang, Feng Liu, Joel Stehouwer, Xiaoming Liu, and Anil~K Jain.
\newblock On the detection of digital face manipulation.
\newblock In {\em CVPR}, 2020.

\bibitem{unified-detection-of-digital-and-physical-face-attacks}
Debayan Deb, Xiaoming Liu, and Anil Jain.
\newblock Unified detection of digital and physical face attacks.
\newblock In {\em arXiv preprint arXiv:2104.02156}, 2021.

\bibitem{goodfellow2014generative}
Ian Goodfellow, Jean Pouget-Abadie, Mehdi Mirza, Bing Xu, David Warde-Farley,
  Sherjil Ozair, Aaron Courville, and Yoshua Bengio.
\newblock Generative adversarial nets.
\newblock In {\em NeurIPS}, 2014.

\bibitem{goodfellow2014explaining}
Ian~J Goodfellow, Jonathon Shlens, and Christian Szegedy.
\newblock Explaining and harnessing adversarial examples.
\newblock In {\em ICLR}, 2015.

\bibitem{huang2018munit}
Xun Huang, Ming-Yu Liu, Serge Belongie, and Jan Kautz.
\newblock Multimodal unsupervised image-to-image translation.
\newblock In {\em ECCV}, 2018.

\bibitem{huang2020nonlinear}
Zhi-Jing Huang, Shan Cheng, Li-Hua Gong, and Nan-Run Zhou.
\newblock Nonlinear optical multi-image encryption scheme with two-dimensional
  linear canonical transform.
\newblock {\em Optics and Lasers in Engineering}, 124:105821, 2020.

\bibitem{isola2017pix2pix}
Phillip Isola, Jun-Yan Zhu, Tinghui Zhou, and Alexei~A Efros.
\newblock Image-to-image translation with conditional adversarial networks.
\newblock In {\em CVPR}, 2017.

\bibitem{jiansheng2009digital}
Mei Jiansheng, Li Sukang, and Tan Xiaomei.
\newblock A digital watermarking algorithm based on {DCT} and {DWT}.
\newblock In {\em WISA}, 2009.

\bibitem{karras2018progressive}
Tero Karras, Timo Aila, Samuli Laine, and Jaakko Lehtinen.
\newblock Progressive growing of {GAN}s for improved quality, stability, and
  variation.
\newblock In {\em ICLR}, 2018.

\bibitem{khan2013digital}
Mohammad~Ibrahim Khan, Md~Maklachur Rahman, and Md~Iqbal~Hasan Sarker.
\newblock Digital watermarking for image authentication based on combined
  {DCT}, {DWT} and {SVD} transformation.
\newblock {\em International Journal of Computer Science Issues}, 10:223, 2013.

\bibitem{liu2019stgan}
Ming Liu, Yukang Ding, Min Xia, Xiao Liu, Errui Ding, Wangmeng Zuo, and Shilei
  Wen.
\newblock {STGAN}: A unified selective transfer network for arbitrary image
  attribute editing.
\newblock In {\em CVPR}, 2019.

\bibitem{liu2018unit}
Ming-Yu Liu, Thomas Breuel, and Jan Kautz.
\newblock Unsupervised image-to-image translation networks.
\newblock In {\em NeurIPS}, 2017.

\bibitem{pscc-net-progressive-spatio-channel-correlation-network-for-image-manipulation-detection-and-localization}
Xiaohong Liu, Yaojie Liu, Jun Chen, and Xiaoming Liu.
\newblock {PSCC-N}et: Progressive spatio-channel correlation network for image
  manipulation detection and localization.
\newblock In {\em arXiv preprint arXiv:2103.10596}, 2021.

\bibitem{liu2018large}
Ziwei Liu, Ping Luo, Xiaogang Wang, and Xiaoou Tang.
\newblock Deep learning face attributes in the wild.
\newblock In {\em ICCV}, 2015.

\bibitem{madry2018towards}
Aleksander Madry, Aleksandar Makelov, Ludwig Schmidt, Dimitris Tsipras, and
  Adrian Vladu.
\newblock Towards deep learning models resistant to adversarial attacks.
\newblock In {\em ICLR}, 2018.

\bibitem{masi2020two}
Iacopo Masi, Aditya Killekar, Royston~Marian Mascarenhas, Shenoy~Pratik
  Gurudatt, and Wael AbdAlmageed.
\newblock Two-branch recurrent network for isolating deepfakes in videos.
\newblock In {\em ECCV}, 2020.

\bibitem{most-gan-3d-morphable-stylegan-for-disentangled-face-image-manipulation}
Safa~C. Medin, Bernhard Egger, Anoop Cherian, Ye Wang, Joshua~B. Tenenbaum,
  Xiaoming Liu, and Tim~K. Marks.
\newblock {MOST-GAN}: 3{D} morphable {S}tyle{GAN} for disentangled face image
  manipulation.
\newblock In {\em AAAI}, 2022.

\bibitem{nataraj2019detecting}
Lakshmanan Nataraj, Tajuddin~Manhar Mohammed, BS Manjunath, Shivkumar
  Chandrasekaran, Arjuna Flenner, Jawadul~H Bappy, and Amit~K Roy-Chowdhury.
\newblock Detecting {GAN} generated fake images using co-occurrence matrices.
\newblock {\em Electronic Imaging}, 2019:532--1, 2019.

\bibitem{nirkin2021deepfake}
Yuval Nirkin, Lior Wolf, Yosi Keller, and Tal Hassner.
\newblock Deepfake detection based on discrepancies between faces and their
  context.
\newblock {\em IEEE Transactions on Pattern Analysis and Machine Intelligence},
  PP:1--1, 2021.

\bibitem{nizan2020council}
Ori Nizan and Ayellet Tal.
\newblock Breaking the cycle - colleagues are all you need.
\newblock In {\em CVPR}, 2020.

\bibitem{park2019gaugan}
Taesung Park, Ming-Yu Liu, Ting-Chun Wang, and Jun-Yan Zhu.
\newblock Gau{GAN}: semantic image synthesis with spatially adaptive
  normalization.
\newblock In {\em ACM}, 2019.

\bibitem{pathakCVPR16context}
Deepak Pathak, Philipp Kr\"ahenb\"uhl, Jeff Donahue, Trevor Darrell, and Alexei
  Efros.
\newblock Context encoders: Feature learning by inpainting.
\newblock In {\em CVPR}, 2016.

\bibitem{pidhorskyi2020alae}
Stanislav Pidhorskyi, Donald~A Adjeroh, and Gianfranco Doretto.
\newblock Adversarial latent autoencoders.
\newblock In {\em CVPR}, 2020.

\bibitem{Pumarola_ijcv2019gananime}
Albert Pumarola, Antonio Agudo, Aleix~M Martinez, Alberto Sanfeliu, and
  Francesc Moreno-Noguer.
\newblock {GAN}imation: One-shot anatomically consistent facial animation.
\newblock {\em International Journal of Computer Vision}, 128:698--713, 2020.

\bibitem{qian2020thinking}
Yuyang Qian, Guojun Yin, Lu Sheng, Zixuan Chen, and Jing Shao.
\newblock Thinking in frequency: Face forgery detection by mining
  frequency-aware clues.
\newblock In {\em ECCV}, 2020.

\bibitem{zhang2021multi}
Zhang Qiu-yu, Jitian Han, and Yutong Ye.
\newblock Multi‐image encryption algorithm based on image hash, bit‐plane
  decomposition and dynamic {DNA} coding.
\newblock {\em IET Image Processing}, 15:885--896, 2020.

\bibitem{quan2018distinguishing}
Weize Quan, Kai Wang, Dong-Ming Yan, and Xiaopeng Zhang.
\newblock Distinguishing between natural and computer-generated images using
  convolutional neural networks.
\newblock {\em IEEE Transactions on Information Forensics and Security},
  13:2772--2787, 2018.

\bibitem{Richter_2016_ECCV}
Stephan~R. Richter, Vibhav Vineet, Stefan Roth, and Vladlen Koltun.
\newblock Playing for data: {G}round truth from computer games.
\newblock In {\em ECCV}, 2016.

\bibitem{rossler2019faceforensics}
Andreas Rossler, Davide Cozzolino, Luisa Verdoliva, Christian Riess, Justus
  Thies, and Matthias Nie{\ss}ner.
\newblock Faceforensics++: Learning to detect manipulated facial images.
\newblock In {\em CVPR}, 2019.

\bibitem{ruiz2020disrupting}
Nataniel Ruiz, Sarah~Adel Bargal, and Stan Sclaroff.
\newblock Disrupting deepfakes: Adversarial attacks against conditional image
  translation networks and facial manipulation systems.
\newblock In {\em ECCV}, 2020.

\bibitem{segalis2020ogan}
Eran Segalis and Eran Galili.
\newblock {OGAN}: Disrupting deepfakes with an adversarial attack that survives
  training.
\newblock {\em arXiv preprint arXiv:2006.12247}, 2020.

\bibitem{singh2012novel}
Amit~Kumar Singh, Nomit Sharma, Mayank Dave, and Anand Mohan.
\newblock A novel technique for digital image watermarking in spatial domain.
\newblock In {\em PDGC}, 2012.

\bibitem{tancik2020stegastamp}
Matthew Tancik, Ben Mildenhall, and Ren Ng.
\newblock Stega{S}tamp: Invisible hyperlinks in physical photographs.
\newblock In {\em CVPR}, 2020.

\bibitem{disentangled-representation-learning-gan-for-pose-invariant-face-recognition}
Luan Tran, Xi Yin, and Xiaoming Liu.
\newblock Disentangled representation learning {GAN} for pose-invariant face
  recognition.
\newblock In {\em CVPR}, 2017.

\bibitem{Tylecek13}
Radim Tyle{\v c}ek and Radim {\v S}{\' a}ra.
\newblock Spatial pattern templates for recognition of objects with regular
  structure.
\newblock In {\em GCPR}, 2013.

\bibitem{wang2021faketagger}
Run Wang, Felix Juefei-Xu, Meng Luo, Yang Liu, and Lina Wang.
\newblock Fake{T}agger: Robust safeguards against deepfake dissemination via
  provenance tracking.
\newblock In {\em ACMM}, 2021.

\bibitem{wang2020fakespotter}
Run Wang, Felix Juefei-Xu, Lei Ma, Xiaofei Xie, Yihao Huang, Jian Wang, and
  Yang Liu.
\newblock Fake{S}potter: A simple yet robust baseline for spotting
  ai-synthesized fake faces.
\newblock In {\em IJCAI}, 2020.

\bibitem{wang2020cnn}
Sheng-Yu Wang, Oliver Wang, Richard Zhang, Andrew Owens, and Alexei~A Efros.
\newblock {CNN}-generated images are surprisingly easy to spot... for now.
\newblock In {\em CVPR}, 2020.

\bibitem{wang2008face}
Xiaogang Wang and Xiaoou Tang.
\newblock Face photo-sketch synthesis and recognition.
\newblock {\em IEEE transactions on pattern analysis and machine intelligence},
  31:1955--1967, 2008.

\bibitem{wang2021realesrgan}
Xintao Wang, Liangbin Xie, Chao Dong, and Ying Shan.
\newblock Real-{ESRGAN}: Training real-world blind super-resolution with pure
  synthetic data.
\newblock In {\em CVPR}, 2021.

\bibitem{wu2020sstnet}
Xi Wu, Zhen Xie, YuTao Gao, and Yu Xiao.
\newblock {SSTNET}: Detecting manipulated faces through spatial, steganalysis
  and temporal features.
\newblock In {\em ICASSP}, 2020.

\bibitem{yavuz2007improved}
Erkan Yavuz and Ziya Telatar.
\newblock Improved {SVD-DWT} based digital image watermarking against watermark
  ambiguity.
\newblock In {\em SAC}, 2007.

\bibitem{ye2020multi}
Huo-Sheng Ye, Nan-Run Zhou, and Li-Hua Gong.
\newblock Multi-image compression-encryption scheme based on quaternion
  discrete fractional hartley transform and improved pixel adaptive diffusion.
\newblock {\em Signal Processing}, 175:107652, 2020.

\bibitem{yeh2020disrupting}
Chin-Yuan Yeh, Hsi-Wen Chen, Shang-Lun Tsai, and Sheng-De Wang.
\newblock Disrupting image-translation-based deepfake algorithms with
  adversarial attacks.
\newblock In {\em WACVW}, 2020.

\bibitem{yi2018dualgan}
Zili Yi, Hao Zhang, Ping Tan, and Minglun Gong.
\newblock Dual{GAN}: Unsupervised dual learning for image-to-image translation.
\newblock In {\em CVPR}, 2017.

\bibitem{finegrained}
A. Yu and K. Grauman.
\newblock Fine-grained visual comparisons with local learning.
\newblock In {\em CVPR}, 2014.

\bibitem{semjitter}
A. Yu and K. Grauman.
\newblock Semantic jitter: Dense supervision for visual comparisons via
  synthetic images.
\newblock In {\em ICCV}, 2017.

\bibitem{yu15lsun}
Fisher Yu, Yinda Zhang, Shuran Song, Ari Seff, and Jianxiong Xiao.
\newblock {LSUN}: Construction of a large-scale image dataset using deep
  learning with humans in the loop.
\newblock {\em arXiv preprint arXiv:1506.03365}, 2015.

\bibitem{zhang2019detecting}
Xu Zhang, Svebor Karaman, and Shih-Fu Chang.
\newblock Detecting and simulating artifacts in {GAN} fake images.
\newblock In {\em WIFS}, 2019.

\bibitem{zhu2018hidden}
Jiren Zhu, Russell Kaplan, Justin Johnson, and Li Fei-Fei.
\newblock {HiDDeN}: Hiding data with deep networks.
\newblock In {\em ECCV}, 2018.

\bibitem{CycleGAN2017}
Jun-Yan Zhu, Taesung Park, Phillip Isola, and Alexei~A Efros.
\newblock Unpaired image-to-image translation using cycle-consistent
  adversarial networks.
\newblock In {\em ICCV}, 2017.

\bibitem{zhu2017bicycle}
Jun-Yan Zhu, Richard Zhang, Deepak Pathak, Trevor Darrell, Alexei~A Efros,
  Oliver Wang, and Eli Shechtman.
\newblock Toward multimodal image-to-image translation.
\newblock In {\em NeurIPS}, 2017.

\bibitem{Zhu_2020_sean}
Peihao Zhu, Rameen Abdal, Yipeng Qin, and Peter Wonka.
\newblock {SEAN}: Image synthesis with semantic region-adaptive normalization.
\newblock In {\em CVPR}, 2020.

\end{thebibliography}
}

\clearpage
\setcounter{equation}{0}
\setcounter{figure}{0}
\setcounter{table}{0}
\setcounter{section}{0}
\twocolumn[\centering \section*{\Large \textbf{Proactive Image Manipulation Detection \\ -- Supplementary material --\\[1cm]}}] 

\Section{Cross Encoder-Template Set Evaluation}
Our framework encrypts a real image using a template from the template set. This encryption would aid in the image manipulation detection if the image is corrupted by any unseen GM. The framework is divided in two stages namely, image encryption and recovery of template  where each stage works independently in inference. 
We therefore provide an ablation to study the performance using different encoder and template set, \ie, we evaluate recovering ability of an encoder using a template set trained with different initialization seeds. The results are shown in Tab.~\ref{tab:diff_seeds}. We observe that even though the template set and the encoder are initialized with different seeds, the performance of our framework doesn't vary much. This shows the stability of our framework even though the initialization seeds of both stages during training are different. 

\begin{figure*}[t!]
\centering
\includegraphics[width=1\textwidth]{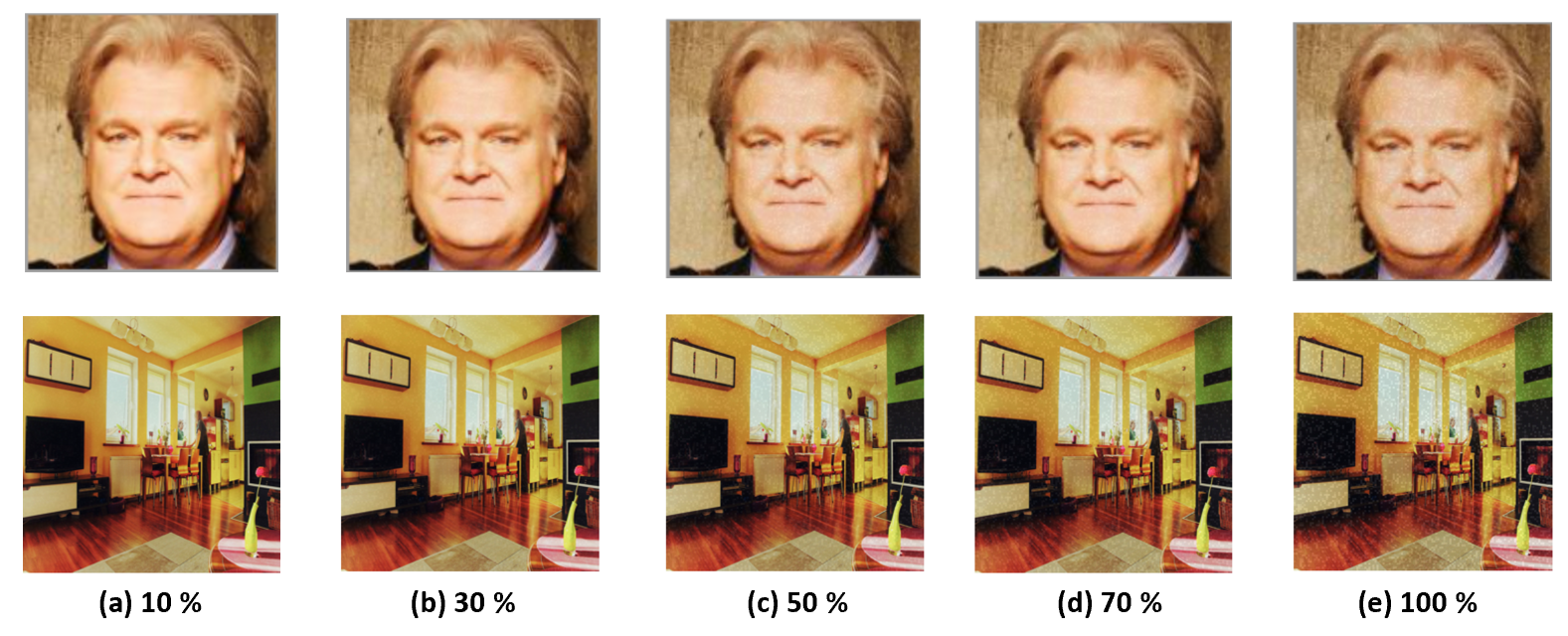}
\caption{Visualization of input images with different template strength. As the template strength is increased, the images become noisier.  }
\label{fig:sig_str_vis}
\end{figure*}

\Section{Template Strength}
We provide the ablation for hyperparameter {\it m} used to control the strength of the added template in Sec.~$4.3$. We observe that the performance is better if we increase the template strength. However, this comes at a trade-off with PSNR which declines if the template strength increases. This is also justified in Fig.~\ref{fig:sig_str_vis} which shows the images with different strength of added template. The images become noisier as the template strength is increased. This is not desirable as there shouldn't be much distortion in the encrypted real image due to our added template. Therefore for our experiments, we select $30\%$ as the strength for the added template.

\Section{Implementation Details}
\minisection{Image editing techniques} We use various image editing techniques in Sec. $4.2$. All the techniques are applied after addition of our template. We provide the implementation details for all these techniques below:
\begin{enumerate}
    \item  Blur: We apply Gaussian blur to the image with 50\% probability using $\sigma$ sampled from $[0, 3]$,
    \item JPEG: We JPEG-compress the image with 50\% probability images using Imaging Library (PIL), with quality sampled from $ \text{Uniform}\{30, 31, . . . , 100\}$.
    \item Blur $+$ JPEG (p): The image is possibly blurred and JPEG-compressed, each with probability p. 
    \item Resizing: We perform the training using $50\%$ of the images with $256 \times 256 \times 3$ resolution and rest with $128 \times 128 \times 3$ resolution images in CelebA-HQ dataset. 
    \item Crop: We randomly crop the images with 50\% probability on each side with pixels sampled from $[0, 30]$. The images are resized to $128 \times 128 \times 3$ resolution.
    \item Gaussian noise: We add Gaussian noise with zero mean and unit variance to the images with 50\% probability.
\end{enumerate}

\begin{table}[t]
\centering
\caption{\small Cross encoder-template set evaluation with different initialization seeds.}
\scalebox{0.8}{
\begin{tabular}{c|c|c|c|c}
\hline
\multicolumn{2}{c|}{Initialization seed} & \multicolumn{3}{c}{Test GM Average precision (\%)}\\\hline
 Encoder& Template set & StarGAN & CycleGAN & GauGAN \\\hline\hline
\multirow{3}{*}{$1$} & $1$ &$96.12$&$100$&$91.62$\\\cline{2-5}
& $2$ &$94.65$&$100$&$91.15$\\\cline{2-5}
& $3$ &$94.83$&$100$&$91.46$\\\hline
\multirow{3}{*}{$2$} & $1$ &$95.48$&$100$&$91.56$\\\cline{2-5}
& $2$ &$95.54$&$100$&$90.85$\\\cline{2-5}
& $3$ &$95.84$&$100$&$91.06$\\\hline
\multirow{3}{*}{$3$} & $1$ &$95.56$&$100$&$91.32$\\\cline{2-5}
& $2$ &$95.62$&$100$&$91.42$\\\cline{2-5}
& $3$ &$96.14$&$100$&$90.41$\\\hline

\hline\hline
\end{tabular}
}
\label{tab:diff_seeds}
\end{table}

\begin{figure*}[!htp]
\centering
\includegraphics[width=\textwidth,,keepaspectratio]{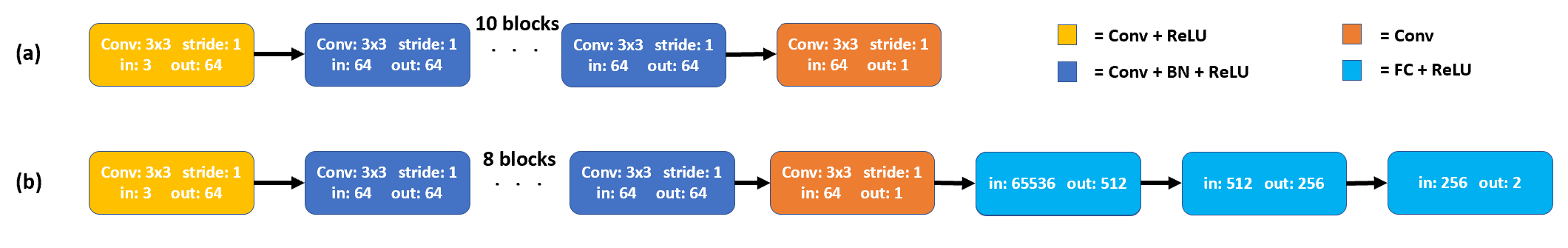}
\caption{Network architecture for our (a) encoder (b) classifier network for image manipulation detection.}
\label{fig:arch}
\end{figure*}

\begin{table*}[t!]
\centering
\caption{\small List of GMs with their datasets and input image resolution used for evaluating our framework's generalization ability. }
\scalebox{0.68}{
\begin{tabular}{c|c|c|c|c|c|c|c|c}
\hline
GM & STGAN~\cite{liu2019stgan} & StarGAN~\cite{choi2018stargan} & CycleGAN~\cite{CycleGAN2017} & GauGAN~\cite{park2019gaugan}& UNIT~\cite{liu2018unit}& MUNIT~\cite{huang2018munit} & StarGAN2~\cite{stargan2} & BicycleGAN~\cite{zhu2017bicycle}\\\hline 
Dataset &CelebA-HQ~\cite{karras2018progressive}&CelebA-HQ~\cite{karras2018progressive}&Facades~\cite{Tylecek13}&COCO~\cite{coco} & GTA$2$City~\cite{Richter_2016_ECCV} & Edges$2$Shoes~\cite{finegrained, semjitter} & CelebA-HQ~\cite{karras2018progressive} & Facades~\cite{Tylecek13}\\\hline
Resolution & $128 \times 128 \times 3$& $256 \times 256 \times 3$& $256 \times 256 \times 3$& $256 \times 256 \times 3$& $512 \times 931 \times 3$& $256 \times 512 \times 3$& $256 \times 256 \times 3$& $256 \times 256 \times 3$\\\hline\hline \\\hline 
GM & CONT\_Encoder~\cite{pathakCVPR16context}  & SEAN~\cite{Zhu_2020_sean} & ALAE\cite{pidhorskyi2020alae} & Pix2Pix\cite{isola2017pix2pix} & DualGAN\cite{yi2018dualgan} & CouncilGAN\cite{nizan2020council} & ESRGAN\cite{wang2021realesrgan} & GANimation\cite{Pumarola_ijcv2019gananime}\\\hline 
Dataset & Paris Street-View~\cite{nizan2020council} & CelebA-HQ~\cite{karras2018progressive} & CelebA-HQ~\cite{karras2018progressive} & Facades~\cite{Tylecek13} & Sketch-Photo~\cite{wang2008face} & CelebA~\cite{liu2018large} & CelebA~\cite{liu2018large} & CelebA~\cite{liu2018large}\\\hline
Resolution & $64 \times 64 \times 3$& $256 \times 256 \times 3$& $256 \times 256 \times 3$& $256 \times 256 \times 3$& $256 \times 256 \times 3$& $256 \times 256 \times 3$& $128 \times 128 \times 3$& $128 \times 128 \times 3$\\\hline\hline
\end{tabular}
}
\label{tab:gm_list}
\end{table*}

\minisection{Network architecture}
Fig.~\ref{fig:arch} shows the network architecture used in different experiments for our framework's evaluation. For our framework, our encoder has $2$ stem convolution layers and $10$ convolution blocks  to recover the added template from encrypted real images. Each block comprises of convolution, batch normalization and ReLU activation. 

In ablation experiments for Table~$8$, we use a classification network with the similar number of layers as our encoder. This is done to show the importance of recovering templates using encoder. This classification networks has $8$ convolution blocks followed by three fully connected layers with ReLU activation in between the layers. The network outputs $2$ dimension logits used for image manipulation detection. 

\begin{figure*}[!htp]
\centering
\includegraphics[width=0.73\textwidth,,keepaspectratio]{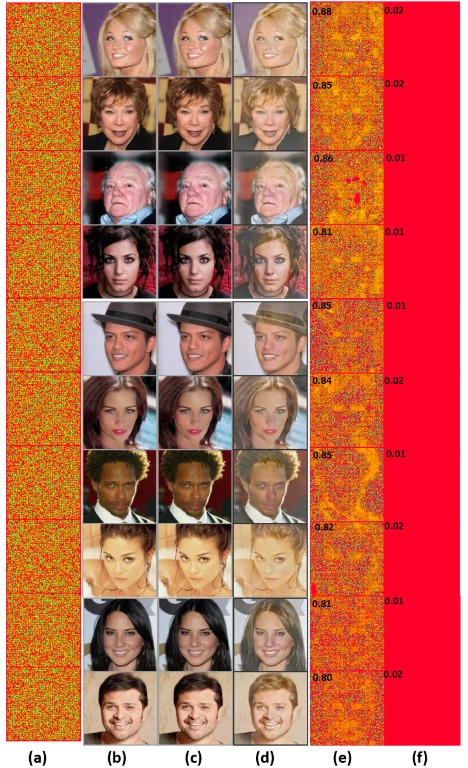}
\caption{Visualization of samples used for GM  STGAN; (a) added template, (b) real images, (c) encrypted real images after adding a template, (d) manipulated images output by a GM, (e) recovered template from (c), and (f) recovered template from (d). Top left corner in last two columns shows the cosine similarity of the recovered template with the added template.}
\label{fig:stgan}
\end{figure*}

\begin{figure*}[!htp]
\centering
\includegraphics[width=0.7\textwidth,,keepaspectratio]{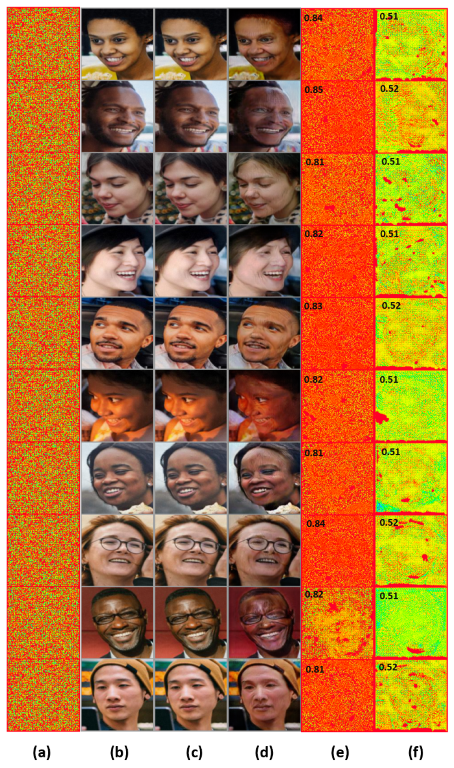}
\caption{Visualization of samples used for GM  StarGAN; (a) added template, (b) real images, (c) encrypted real images after adding a template, (d) manipulated images output by a GM, (e) recovered template from (c), and (f) recovered template from (d). Top left corner in last two columns shows the cosine similarity of the recovered template with the added template.}
\label{fig:stargan}
\end{figure*}

\begin{figure*}[!htp]
\centering
\includegraphics[width=0.7\textwidth,keepaspectratio]{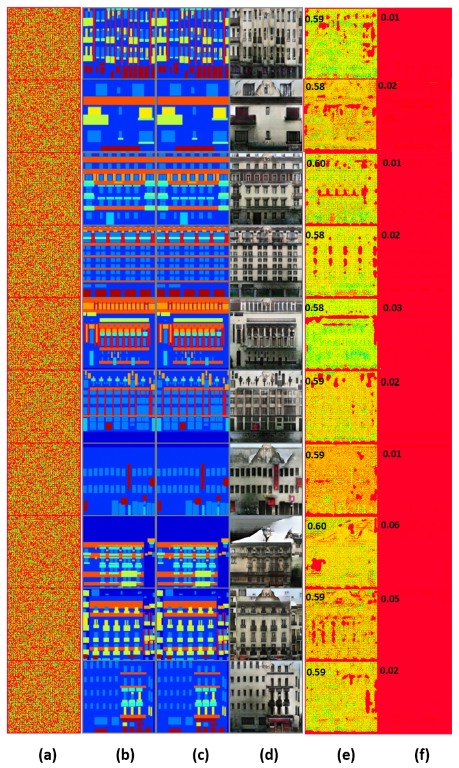}
\caption{Visualization of samples used for GM  CycleGAN; (a) added template, (b) real images, (c) encrypted real images after adding a template, (d) manipulated images output by a GM, (e) recovered template from (c), and (f) recovered template from (d). Top left corner in last two columns shows the cosine similarity of the recovered template with the added template.}
\label{fig:cyclegan}
\end{figure*}

\begin{figure*}[!htp]
\centering
\includegraphics[width=0.7\textwidth,,keepaspectratio]{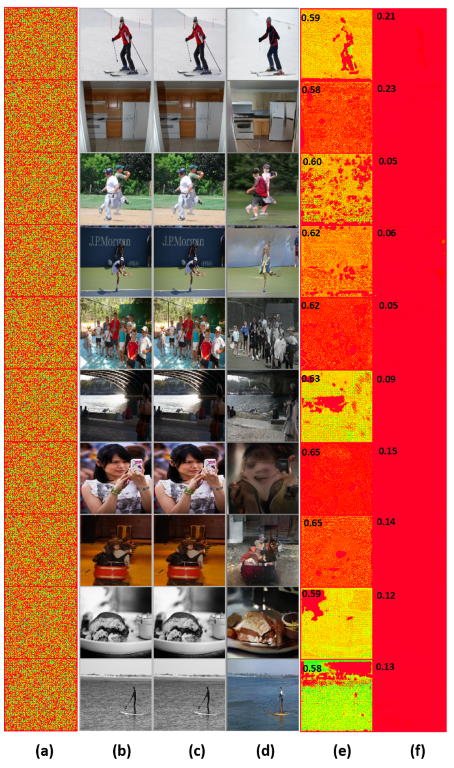}
\caption{Visualization of samples used for GM  GauGAN; (a) added template, (b) real images, (c) encrypted real images after adding a template, (d) manipulated images output by a GM, (e) recovered template from (c), and (f) recovered template from (d). Top left corner in last two columns shows the cosine similarity of the recovered template with the added template.}
\label{fig:gaugan}
\end{figure*}

\begin{figure*}[!htp]
\centering
\includegraphics[height=0.43\textheight,keepaspectratio]{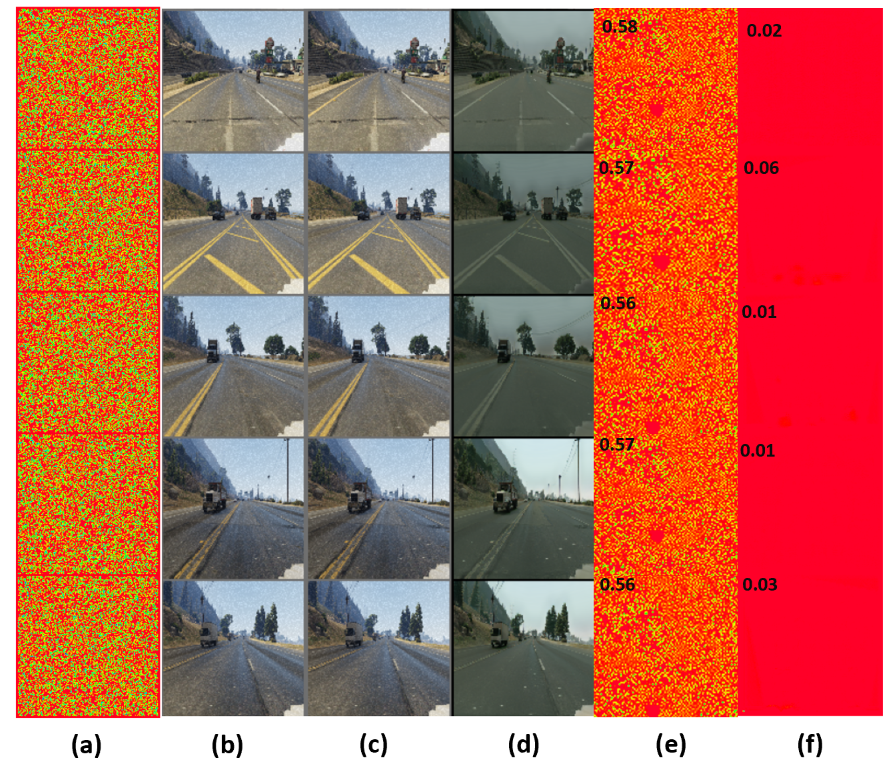}
\caption{Visualization of samples used for GM  UNIT; (a) added template, (b) real images, (c) encrypted real images after adding a template, (d) manipulated images output by a GM, (e) recovered template from (c), and (f) recovered template from (d). Top left corner in last two columns shows the cosine similarity of the recovered template with the added template.}
\label{fig:unit}
\end{figure*}

\begin{figure*}[!htp]
\centering
\includegraphics[height=0.43\textheight,,keepaspectratio]{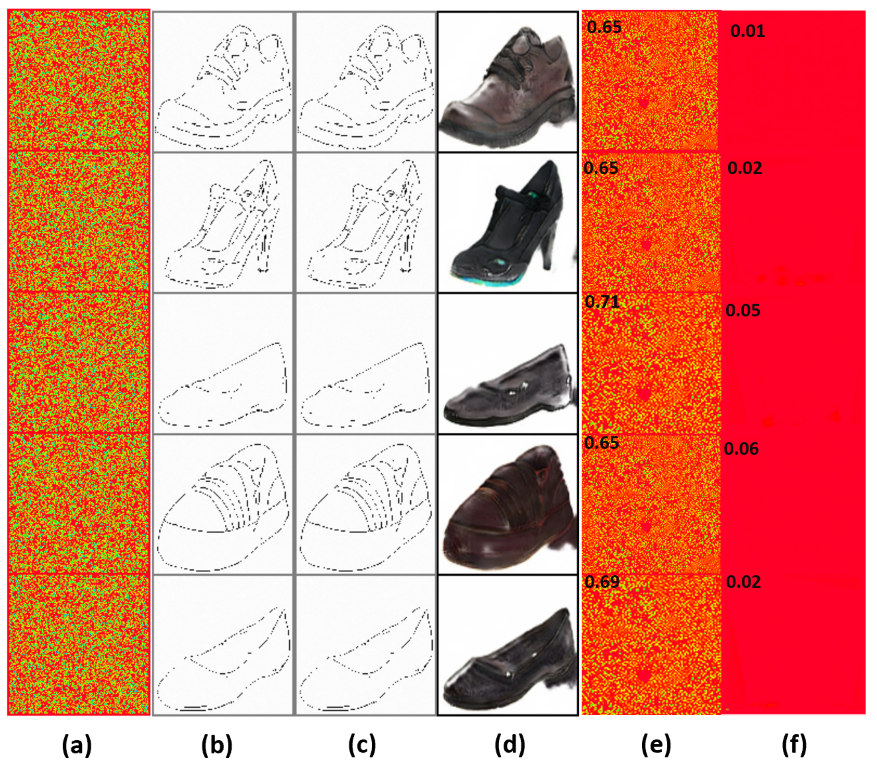}
\caption{Visualization of samples used for GM  MUNIT; (a) added template, (b) real images, (c) encrypted real images after adding a template, (d) manipulated images output by a GM, (e) recovered template from (c), and (f) recovered template from (d). Top left corner in last two columns shows the cosine similarity of the recovered template with the added template.}
\label{fig:munit}
\end{figure*}

\begin{figure*}[!htp]
\centering
\includegraphics[height=0.43\textheight,,keepaspectratio]{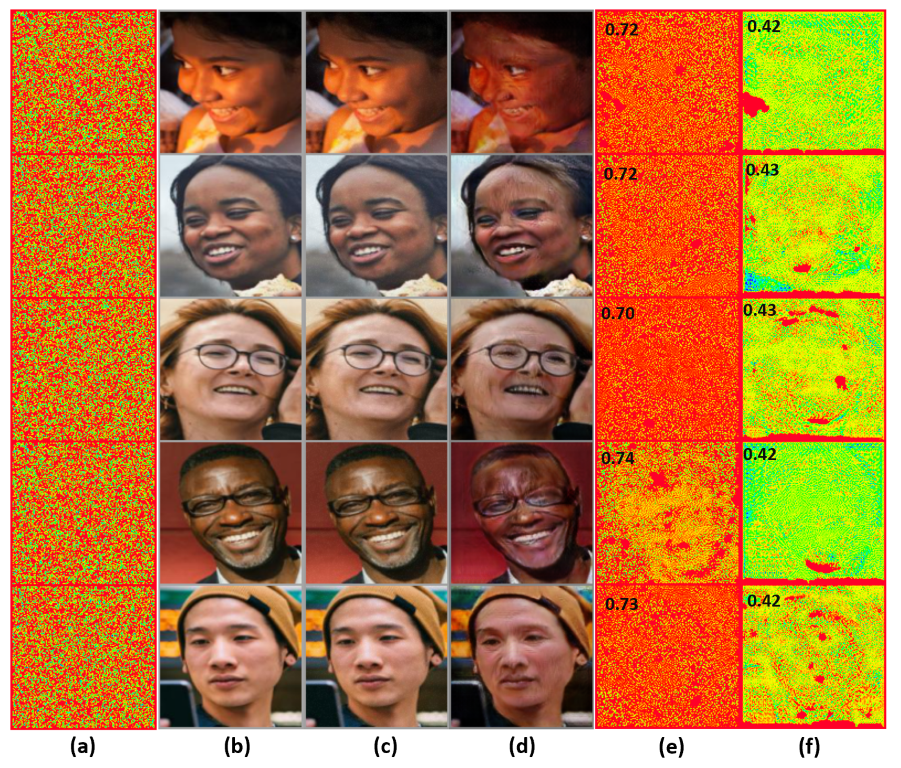}
\caption{Visualization of samples used for GM  StarGANv2; (a) added template, (b) real images, (c) encrypted real images after adding a template, (d) manipulated images output by a GM, (e) recovered template from (c), and (f) recovered template from (d). Top left corner in last two columns shows the cosine similarity of the recovered template with the added template.}
\label{fig:stargan2}
\end{figure*}

\begin{figure*}[!htp]
\centering
\includegraphics[height=0.43\textheight,keepaspectratio]{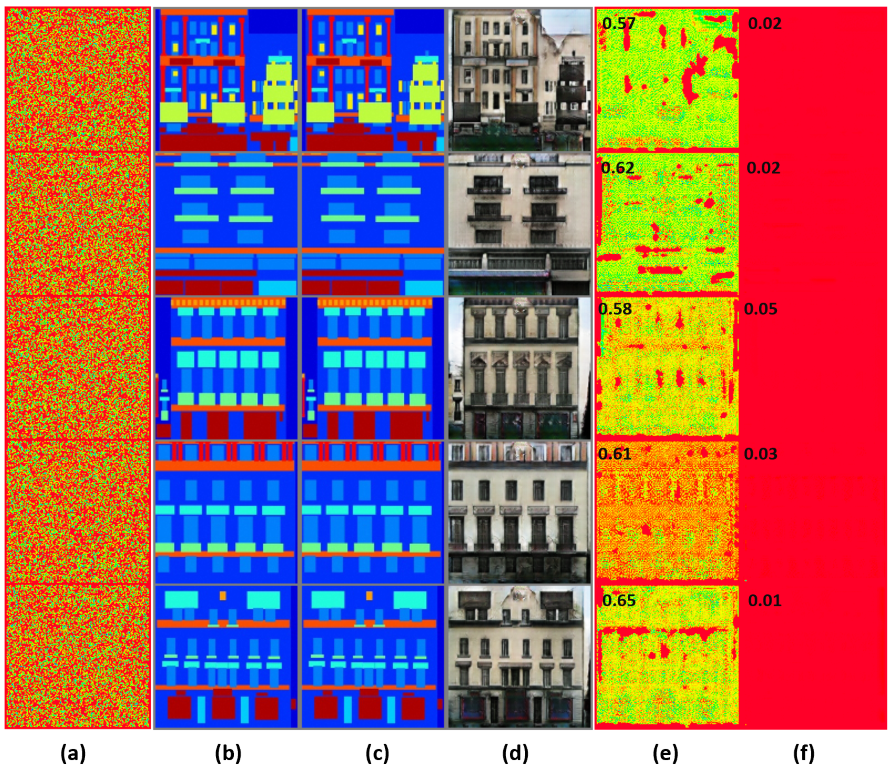}
\caption{Visualization of samples used for GM  BicycleGAN; (a) added template, (b) real images, (c) encrypted real images after adding a template, (d) manipulated images output by a GM, (e) recovered template from (c), and (f) recovered template from (d). Top left corner in last two columns shows the cosine similarity of the recovered template with the added template.}
\label{fig:bicyclegan}
\end{figure*}

\begin{figure*}[!htp]
\centering
\includegraphics[height=0.43\textheight,,keepaspectratio]{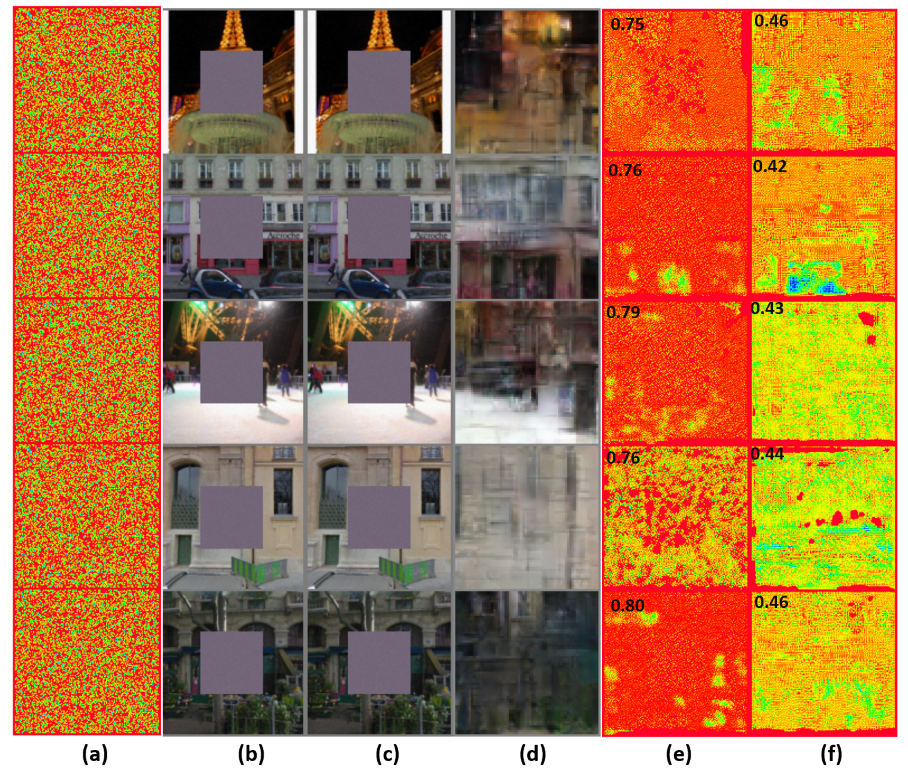}
\caption{Visualization of samples used for GM  CONT\_Encoder; (a) added template, (b) real images, (c) encrypted real images after adding a template, (d) manipulated images output by a GM, (e) recovered template from (c), and (f) recovered template from (d). Top left corner in last two columns shows the cosine similarity of the recovered template with the added template.}
\label{fig:context}
\end{figure*}

\begin{figure*}[!htp]
\centering
\includegraphics[height=0.43\textheight,keepaspectratio]{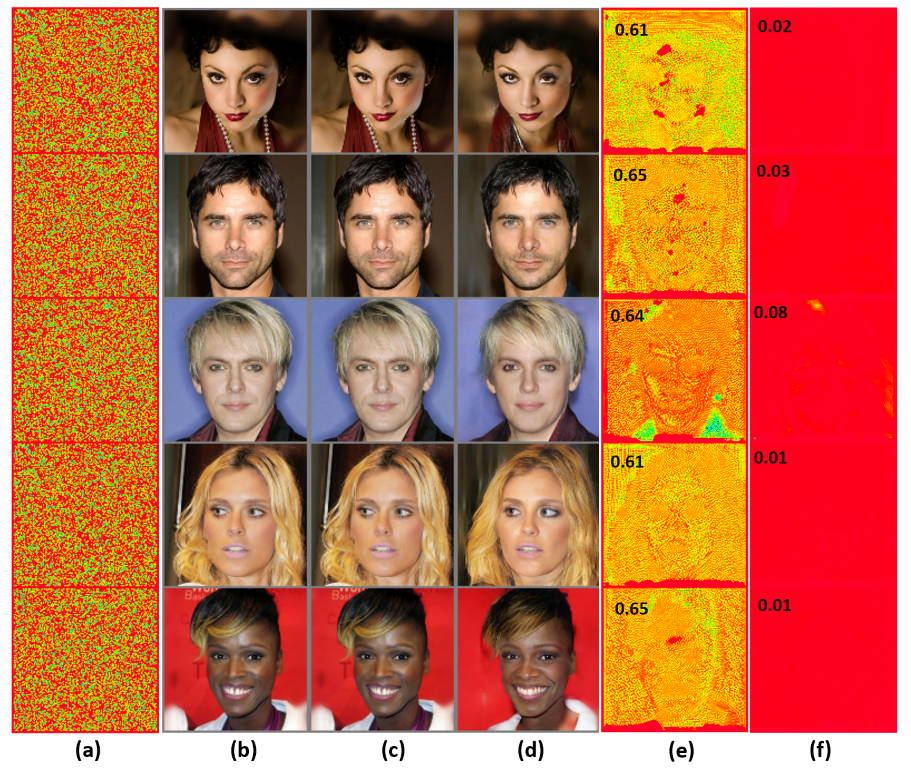}
\caption{Visualization of samples used for GM  SEAN; (a) added template, (b) real images, (c) encrypted real images after adding a template, (d) manipulated images output by a GM, (e) recovered template from (c), and (f) recovered template from (d). Top left corner in last two columns shows the cosine similarity of the recovered template with the added template.}
\label{fig:sean}
\end{figure*}

\begin{figure*}[!htp]
\centering
\includegraphics[height=0.43\textheight,,keepaspectratio]{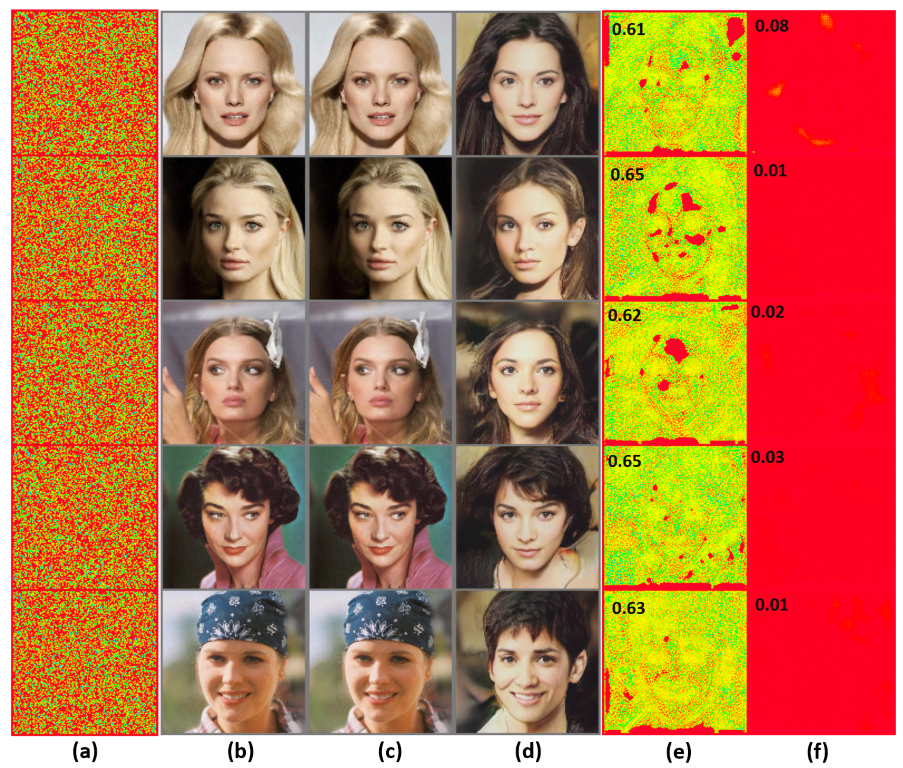}
\caption{Visualization of samples used for GM  ALAE; (a) added template, (b) real images, (c) encrypted real images after adding a template, (d) manipulated images output by a GM, (e) recovered template from (c), and (f) recovered template from (d). Top left corner in last two columns shows the cosine similarity of the recovered template with the added template.}
\label{fig:alae}
\end{figure*}

\begin{figure*}[!htp]
\centering
\includegraphics[height=0.43\textheight,keepaspectratio]{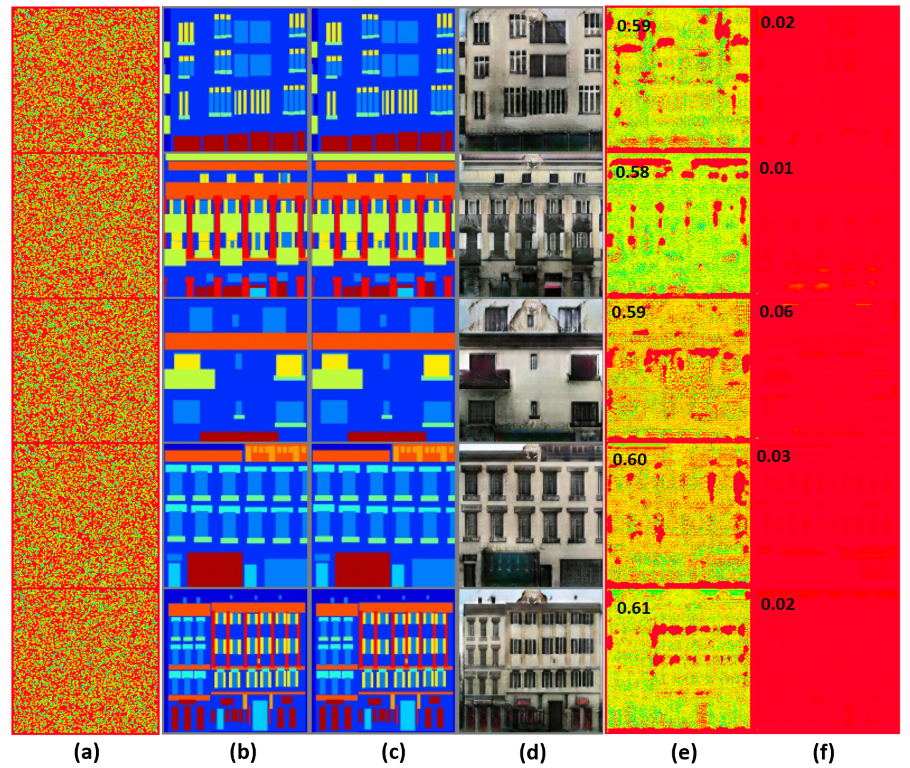}
\caption{Visualization of samples used for GM  Pix2Pix; (a) added template, (b) real images, (c) encrypted real images after adding a template, (d) manipulated images output by a GM, (e) recovered template from (c), and (f) recovered template from (d). Top left corner in last two columns shows the cosine similarity of the recovered template with the added template.}
\label{fig:pix2pix}
\end{figure*}

\begin{figure*}[!htp]
\centering
\includegraphics[height=0.43\textheight,,keepaspectratio]{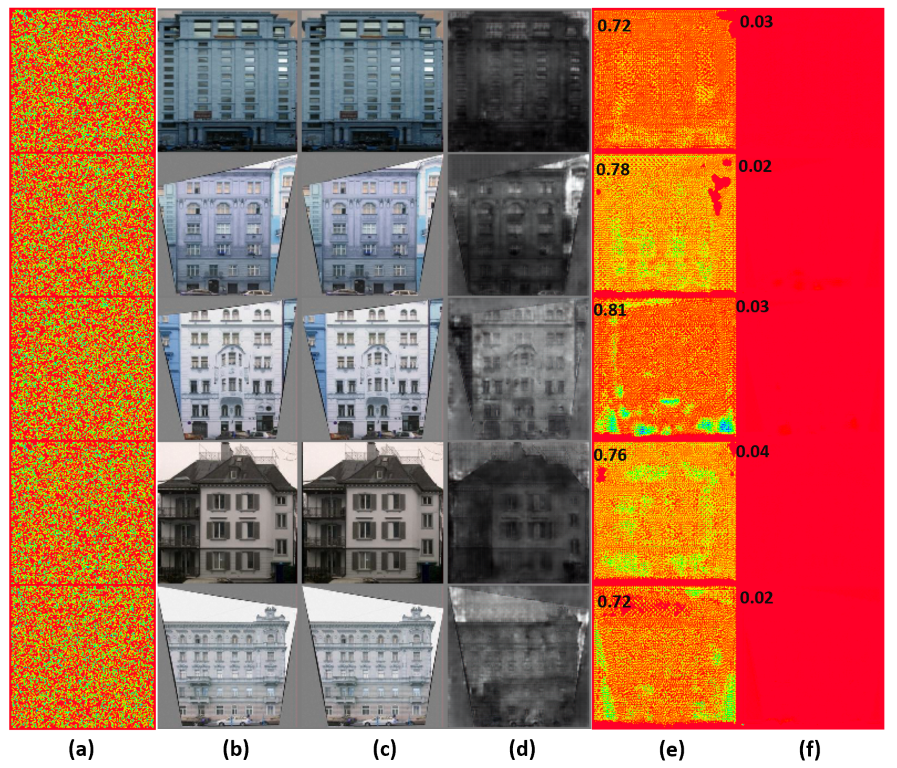}
\caption{Visualization of samples used for GM  DualGAN; (a) added template, (b) real images, (c) encrypted real images after adding a template, (d) manipulated images output by a GM, (e) recovered template from (c), and (f) recovered template from (d). Top left corner in last two columns shows the cosine similarity of the recovered template with the added template.}
\label{fig:dualgan}
\end{figure*}

\begin{figure*}[!htp]
\centering
\includegraphics[height=0.43\textheight,keepaspectratio]{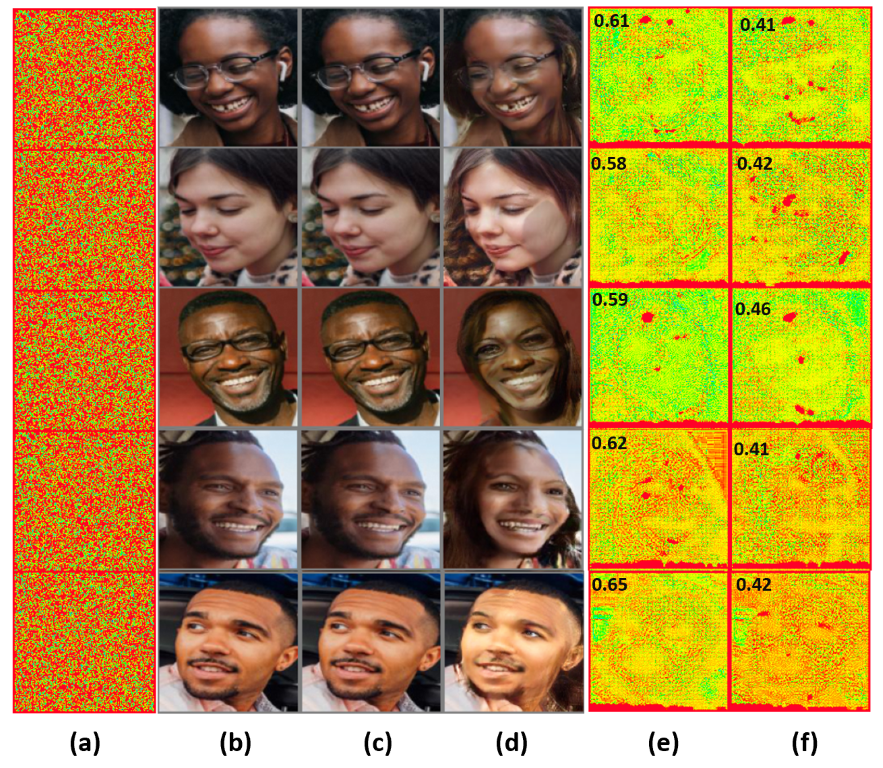}
\caption{Visualization of samples used for GM  CouncilGAN; (a) added template, (b) real images, (c) encrypted real images after adding a template, (d) manipulated images output by a GM, (e) recovered template from (c), and (f) recovered template from (d). Top left corner in last two columns shows the cosine similarity of the recovered template with the added template.}
\label{fig:councilgan}
\end{figure*}

\begin{figure*}[!htp]
\centering
\includegraphics[height=0.43\textheight,,keepaspectratio]{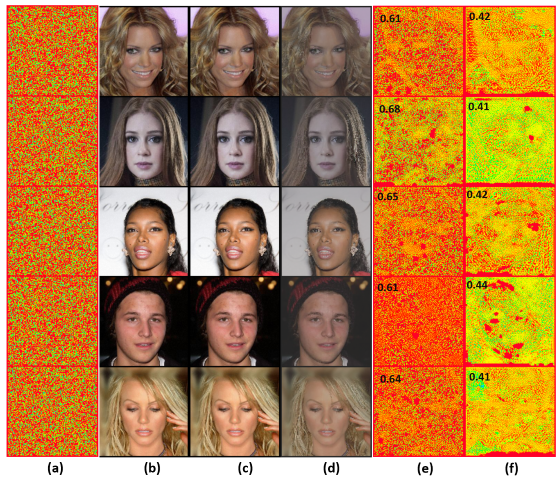}
\caption{Visualization of samples used for GM  ESRGAN; (a) added template, (b) real images, (c) encrypted real images after adding a template, (d) manipulated images output by a GM, (e) recovered template from (c), and (f) recovered template from (d). Top left corner in last two columns shows the cosine similarity of the recovered template with the added template.}
\label{fig:esrgan}
\end{figure*}

\begin{figure*}[!htp]
\centering
\includegraphics[height=0.43\textheight,keepaspectratio]{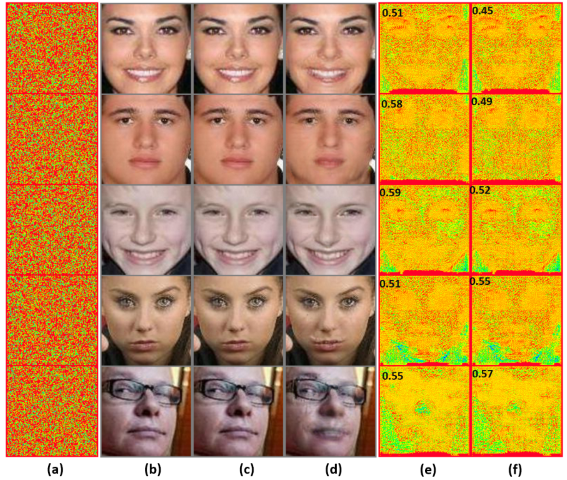}
\caption{Visualization of samples used for GM  GANimation; (a) added template, (b) real images, (c) encrypted real images after adding a template, (d) manipulated images output by a GM, (e) recovered template from (c), and (f) recovered template from (d). Top left corner in last two columns shows the cosine similarity of the recovered template with the added template.}
\label{fig:ganimation}
\end{figure*}

\Section{List of GMs}
We use a variety of GMs to test the generalization ability of our framework. These GMs have varied network architectures and many of them are trained on different datasets. We summarize all the GMs in Tab.~\ref{tab:gm_list}. We also provide visualization for different real image samples used in evaluating the performance for all these GMs in Fig.~\ref{fig:stgan} -~\ref{fig:ganimation}. We show the added template and the recovered templates in ``gist\_rainbow" cmap for better visualization and indicate the cosine similarity of the recovered template with the added template. As shown in Fig.~\ref{fig:stgan} for training with STGAN, the encrypted real images have higher cosine similarity compared to their manipulated counterparts. However, during testing, the difference between the two cosine similarities decreases as shown in Fig.~\ref{fig:stargan} -~\ref{fig:ganimation} for different GMs.

\Section{Dataset License Information}
We use diverse datasets for our experiments which include face and non-face datasets. For face datasets, we use existing datasets including CelebA~\cite{liu2018large} and CelebA-HQ~\cite{karras2018progressive}. The CelebA dataset contains images entirely from the internet and has no associated IRB approval. The authors mention that the dataset is available for non-commercial research purposes only, which we strictly adhere to. We only use the database internally for our work and primarily for evaluation.  CelebA-HQ consists images collected from the internet. Although there is no associated IRB approval, the authors assert in the dataset agreement that the dataset is only to be used for non-commercial research purposes, which we strictly adhere to. 

We use some non-face datasets too for our experiments. The Facades~\cite{Tylecek13} dataset was collected at the Center for Machine Perception and is provided under Attribution-ShareAlike license. Edges2Shoes~\cite{finegrained, semjitter} is a large shoe dataset consisting of images collected from \url{https://www.zappos.com}. The authors mention that this dataset is for academic, non-commercial use only. GTA2City~\cite{Richter_2016_ECCV} dataset consists of a large number of densely labelled frames extracted from computer games. The authors mention that the data is for research and educational use only. The sketch-photo~\cite{wang2008face} datset refers to the CUHK face sketch FERET database. The authors assert in the dataset agreement that the dataset is only to be used for noncommercial research purposes, which we strictly adhere to. Paris street-view~\cite{nizan2020council} dataset contains images collected using google street view and is to be used for noncommercial research purposes. 

\end{document}